\pdfoutput=1
\documentclass{article}

\usepackage[square, comma, sort&compress, numbers]{natbib}

\usepackage[preprint]{nips_2018}

\usepackage{amsmath}
\usepackage[utf8]{inputenc} 
\usepackage[T1]{fontenc}    
\usepackage{hyperref}       
\usepackage{url}            
\usepackage{booktabs}       
\usepackage{amsfonts}       
\usepackage{nicefrac}       
\usepackage{microtype}      
\usepackage{algorithmic}
\usepackage{epsfig}
\usepackage{stmaryrd}
\usepackage{listings}
\usepackage{color}

\DeclareMathOperator*{\argmax}{arg\,max}

\definecolor{codegreen}{rgb}{0,0.55,0.2}
\definecolor{codegray}{rgb}{0.5,0.5,0.5}
\definecolor{codeblue}{rgb}{0.1,0,0.56}
\definecolor{backcolour}{rgb}{1,1,1}
 
\lstdefinestyle{codestyle}{
    backgroundcolor=\color{backcolour},   
    commentstyle=\color{codegreen},
    keywordstyle=\color{codeblue},
    numberstyle=\tiny\color{codegray},
    stringstyle=\color{codegreen},
    basicstyle=\footnotesize,
    breakatwhitespace=false,         
    breaklines=true,                 
    captionpos=b,                    
    keepspaces=true,                 
    numbers=left,                    
    numbersep=5pt,                  
    showspaces=false,                
    showstringspaces=false,
    showtabs=false,                  
    tabsize=2
}
 
\title{DynamicEmbedding: Extending TensorFlow for Colossal-Scale Applications}

\author{
 Yun Zeng \\
 Google \\
 \texttt{xzeng@google.com} \\
 \And
Siqi Zuo \\
Google \\
\texttt{siqiz@google.com} \\
\And
Dongcai Shen \\
Google \\
\texttt{dongcai@google.com}  \\
}

\begin{document}

\maketitle

\begin{abstract}
One of the limitations of deep learning models with sparse features today stems from the predefined nature of their input, which requires a dictionary be defined prior to the training. With this paper we propose both a theory and a working system design which remove this limitation, and show that the resulting models are able to perform better and efficiently run at a much larger scale. Specifically, we achieve this by decoupling a model’s content from its form to tackle architecture evolution and memory growth separately.
To efficiently handle model growth, we propose a new neuron model, called \emph{DynamicCell}, 
drawing inspiration from from the free energy principle~\cite{free_energy_friston06} to introduce the concept of \emph{reaction} to discharge non-digestive energy, which also subsumes gradient descent based approaches as its special cases.
We implement \emph{DynamicCell} by introducing a new server into TensorFlow to take over most of the work involving model growth.
Consequently, it enables any existing deep learning models to efficiently handle arbitrary number of distinct sparse features (\emph{e.g.}, search queries), and grow incessantly without redefining the model.
Most notably, one of our models, which has been reliably running in production for over a year, is capable of suggesting high quality keywords for advertisers of Google Smart Campaigns~\cite{awx} and achieved significant accuracy gains based on a challenging metric -- evidence that data-driven, self-evolving systems can potentially exceed the performance of traditional rule-based approaches.
\end{abstract}

\section{Introduction}

\small{``And while growth is a somewhat vague word for a very complex matter, ... it deserves to be studied in relation to form: whether it proceed by simple increase of size without obvious alteration of form, or whether it so proceed as to bring about a gradual change of form and the slow development of a more or less complicated structure.''}
\\[3pt]
\rightline{{\rm --- D'Arcy Wentworth Thompson in \emph{On Growth and Form}}}
\\[3pt]

\subsection{Motivation}

Big data applications are facing an increasingly large and diverse set of potential inputs for their machine learning models.
Existing libraries for deep learning (\emph{e.g.}, \cite{tensorflow,theano,caffe,pytorch}) mostly focus on minimizing certain loss functions defined by static models, leading to challenges of implementing models that can grow and self-evolve (\emph{e.g.}, ~\cite{progress_network}). 
While there is increasing popularity of the automatic process of searching for optimal neural network architectures~\cite{automl_2016} or model evolutions~\cite{neuroevolution}, the process for a model to grow is usually ignored.
In this paper, we introduce a system called \emph{DynamicEmbedding} that is capable of growing itself by learning from potentially \emph{unlimited} novel input, which could be useful in scenarios where focusing on the most frequently occurring data may still discard too much useful information.

To understand a fundamental limitation among existing designs of deep learning libraries, let us consider a simple example of training a skip-gram model~\cite{emb_mikolov2013word2vec} (\emph{a.k.a.}, Word2Vec) based on new articles that emerge everyday from online newspapers.
The training instances here for the model are tuples of words that are next to each other (cooccurrences), and the desired outcome is a mapping from each word to a vector space (or embedding) in which semantically similar words are close to each other.
A common pre-processing step for implementing the Word2Vec algorithm is to define a \emph{dictionary} of variables that contains all the words whose embeddings should be learned.
The requirement of a dictionary before training is what limits the model to grow, either into handling never seen words or into increasing the dimensionality of the embedding.


\subsection{Core claims}

In an attempt to search for a framework that better accommodates model growth, we start from a recent work~\cite{caml} that treats input/output of a neural network layer as sufficient statistics (embeddings) of certain distribution, and further connect it to the free energy principle~\cite{free_energy_friston06} by proposing a new neuron model called \emph{DynamicCell}.
Intuitively, it allows a neural network layer to minimize its free energy by both regulating internal state and taking actions. 
In addition, when the input contains non-digestive energy, it also discharges it through \emph{reaction}, in order to maintain a stable internal state.
We show that this slight modification to the free-energy principle allows us to connect it to traditional gradient descent based approaches. As a result, an input signal to a layer can be processed either continuously, or combinatorially. For example, when a novel input feature (e.g., a new word) is seen from the input, a layer could dynamically allocate an embedding for it and send it to upstream layers for further processing.

To implement the above ideas, some major changes to TensorFlow~\cite{tensorflow} are needed.
Specifically, a new set of operations are added to TensorFlow's Python API to directly take symbolic strings as input, as well as to ``intercept'' the forward (upstream) and backward (downstream) signals when running a model.
These operations then talk to a new server called DynamicEmbedding Service (DES) to process the content part of a model.
During a forward execution of a model, these operations retrieve the underlying floating point values (embeddings) for their layer input from DES, and pass them to the layer output.
Similarly during backward execution, the computed gradients or any other information, are passed into DES for updating internal states based on algorithms customizable by our users.

In fact, DES plays a key role in expanding the capacity of TensorFlow, reflected in the following aspects:
\begin{itemize}
\item \emph{Virtually unlimited capacity for embedding data}: by collaborating with external storage systems such as Bigtable~\cite{bigtable} or Spanner~\cite{spanner}, it pushes a model's capacity to the limit of storage \footnote{Among our internal communications with multiple teams inside Google, what most engineers are excited about our system is its flexibility in training a model directly on Bigtable, or any other external storage.}. In fact, our system is designed to be modular to work with any storage system that supports key/value data lookup/update.
\item \emph{Flexible gradient descent update}: DES can keep global information about a layer, such as word frequencies or average gradient changes, to help it decide when to update an embedding. Gradient descent update is no longer a homogeneous process for every variable, and each layer can maintain their own ``homeostasis''~\cite{life_book} by taking proper actions\footnote{According to the free-energy principle~\cite{free_energy_friston06}, homeostasis is a biological system's resistance to disorder, carried out by an active inference process to minimize surprises. Our systems allows each layer (neuron) to take its own actions to maintain its ``homeostasis''.}. 
Meanwhile it is also guaranteed that our new system is backward compatible with any gradient descent optimizers (\emph{e.g.}, SGD~\cite{sgd}, AdaGrad~\cite{tf_adagrad}, Momentum~\cite{gradient_descent_momentum}).
\item \emph{Efficiency}: Computational/memory load on DES is automatically distributed into its worker machines in the cloud. Training speed is proportional to the number of Tensorflow workers, and model capacity is decided by the number of DynamicEmbedding workers.
\item \emph{Reliability}: With DES, a TensorFlow model becomes very small as most data are saved to external storage like Bigtable. Therefore training a large model becomes resilient to machine failures that are due to exceeded resources (memory or CPU).
\item \emph{Inherent support for transfer or multitask learning}: By taking the embedding data out of TensorFlow, multiple models can share the same layer, simply by using the same operation name and DES configuration. Therefore, model sharing becomes a norm rather than an option\footnote{Note that TensorFlow's TF Hub component can also achieve similar goal.}.
\end{itemize}

Our DynamicEmbedding system has been proven to be particularly important in large-scale deep learning systems, and has been robustly running in multiple applications for more than one year.
Its implementation guarantees that a TensorFlow model with DynamicEmbedding runs at least as fast as without it (assuming sufficient reliability in the remote storage system), with additional benefits like much bigger capacity, less code to write, and much less data pre-processing work. 
The major overhead for our engineers to switch to DynamicEmbedding is the new APIs to learn and configuration details for external storages like Bigtable or Spanner, which we are trying to simplify as much as possible.

In the past two years since our system was launched, we have migrated many popular models, especially those involving sparse features before training, to enjoy the benefit of continual growth from input.
For example, with upgraded Google Inception model for image annotation~\cite{inception}, it can learn from labels obtained from the enormous search queries;
with upgraded Google Neural Machine Translation (GNMT) model for translation~\cite{gnmt}, it has been applied for translating sentences into ad descriptions, in multiple languages;
our upgraded Word2Vec model now keeps mapping an ever growing repository of search queries into their embedding space, allowing us to quickly find semantically similar queries to any root query, in any language.

By adopting DynamicEmbedding, we find training a single model without much data pre-processing is sufficient to achieve satisfactory performance.
In particular, one of our sparse feature models for suggesting keywords from website content (in any languages) achieved highly accurate results compared to other rule-based systems -- an evidence that by allowing a system to self-evolve driven by data, it is possible to quickly outperform those that are evolved by human tunings.

\begin{figure}[t]
\centering
\begin{tabular}{c}
    \mbox{\epsfig{figure= 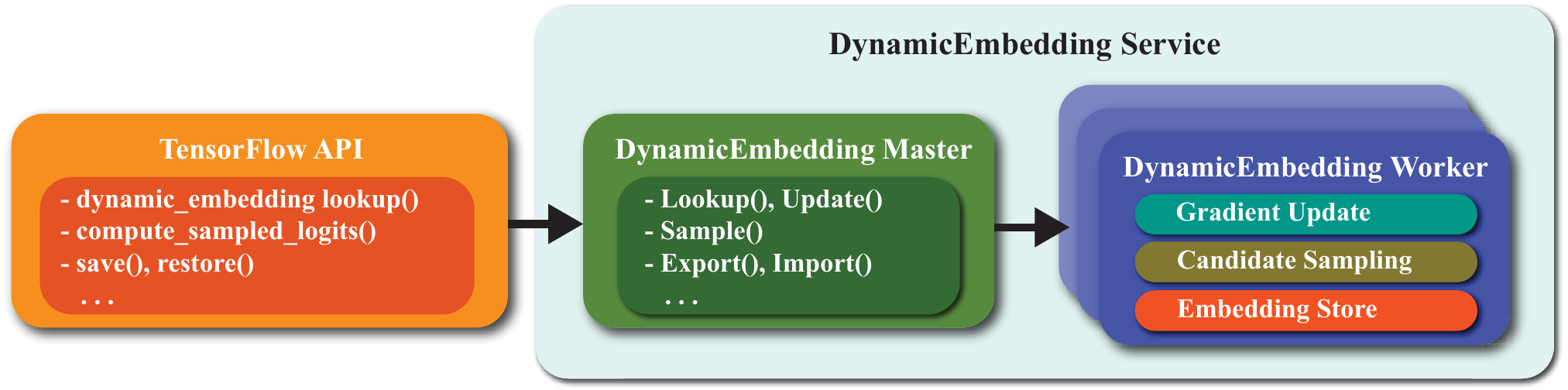, height = 3.2cm}}
\end{tabular}
\caption{\small DynamicEmbedding system overview. It extends the capacity of TensorFlow by introducing a few new APIs to process id-to-dense or dense-to-id operations directly without requiring a dictionary. These APIs talk to DynamicEmbedding Master during forward and backward executions of a TensorFlow computation graph, which in turn divide the work and distribute them into different DynamicEmbedding Workers. } \label{fig:overview}
\end{figure}

\paragraph{System overview}
Fig.~\ref{fig:overview} illustrates an overview of our extended components added to TensorFlow.
The overall goal is to let existing TensorFlow APIs only handle the static (form) part of a model: defining nodes, their connections and passing data between them, and hand the task of trainable variable lookup/update/sample to our DynamicEmbedding Service (DES) to allow them be constructed and grow dynamically.
Besides, we also need to define a new set of Python APIs that directly take string Tensors as input and output their embeddings, or vice versa~\footnote{Previously this needs to be done by first defining floating variables for these string input/output.}.
These TensorFlow APIs directly talk to a component called DynamicEmbedding Master (DEM), which in turn distributes the actual job into DynamicEmbedding Workers (DEWs).
DEWs are responsible for embedding lookup/update/sample, communicating to external cloud storage such as Bigtable or Spanner, updating embedding values based on various gradient descent algorithms to be backward compatible, etc.

\paragraph{Paper organization}
The rest of this paper is organized as follows: we first introduce our new neural network model in Sec.~\ref{section:math_formulation};
The design and implementation details of DynamicEmbedding are discussed next, including its TensorFlow APIs (Sec.~\ref{section:tensorflow_api}) and various components of DynamicEmbedding Service (Sec.~\ref{section::des});
Results from real-world applications are demonstrated in Sec.~\ref{section:results};
Finally, we conclude our work in Sec.~\ref{section:conclusion}.

\section{Mathematical Formulation}
\label{section:math_formulation}

A basic idea in the free energy principle stipulates that a biological system tends to minimize ``surprise'' that is defined as $\log \frac{1}{P(s|m)}$, where $s$ is current internal and external state of a system and $m$ is an internal model that explains $s$.
We can connect such an idea to neural networks, by redefining ``surprise'' as the difference (measured by KL-divergence) between a system's state distribution with and without its contextual input, denoted as $P(w|c)$ and $P(w)$, respectively\footnote{For example, if $c$ denotes very low temperature felt from the environment by a dog, its internal state would change dramatically to protect itself from freezing (a death threat). A big difference between $P(w)$ and $P(w|c)$ would cause a biological system either adjust itself to adapt to the environment (\emph{e.g.}, by growing more fur), or take actions to change the sensory input (\emph{e.g.}, by migrating to a warmer place).}.
Compared to the original formulation mentioned above, our new treatment can be implemented at a cell level, and eliminates the need for an internal, predictive model $m$ to explain state $s$, which itself can be a complex process.
We will show that \emph{the back-propagation algorithm belongs to a general process of free-energy minimization in the embedding space},
which brings a new outlook to the artificial neural network (ANN) as we know it: an ANN is a group of inter-connected neurons minimizing it own free energy\footnote{To some extent, this hints why a life's form can range from prokaryotes (single-celled organism) to complex ones like sapiens -- any number of self-interested, surprising minimizing cells can form a life if they can improve the overall survival rate. Therefore, a corporate like Google, is also a form of life, as suggested in~\cite{yuval_book}.}.
In the rest of this section, we will explain in details our new treatment of neural networks and starting from next section, we will show the practical impact brought by it, \emph{i.e.}, a new system design and improved performance in real-world applications.

\subsection{Exponential family, embedding and artificial neural networks}\label{subsec:exp_family}
The idea of using neural networks to represent the embeddings of sparse features has been widely explored in language related models~\cite{emb_bengio2003neural,emb_mikolov2013word2vec}.
In essence, a layer in a neural network is nothing but a representation of the sufficient statistics of its variables for a certain distribution $P(w_1, \ldots, w_n | c_1, \ldots, c_m)$.
\cite{caml} further generalizes such an idea to connect with many existing deep neural network models by considering their probabilistic forms, and derived a new equation in the embedding space that takes into account the relevance of contextual input to the output.
For example, a layer in a neural network can be regarded as a representation of the distribution $P(w|c)$ in the embedding space, where $c$ is the contextual input to the layer and $w$ is the output.
By further assuming $P(w|c)\propto \exp(\langle \vec{w}, \vec{c}\rangle)$\footnote{From the most general form $P(w|c)\propto \exp(E(\vec{w}, \vec{c}))$ and the Taylor's theorem for multivariable functions~\cite{math_analysis_book}, one can always approximate $P(w|c)$ by $\exp(\langle \vec{w}, \vec{c}\rangle) / Z(\vec{c})$ where $\vec{w}$ may also depend on $\vec{c}$.}, where $\vec{w}$ and $\vec{c}$ represent the embeddings of $w$ and $c$, respectively, then a layer simply computes $\vec{w}$ based on $\vec{c}$. 

This challenges the conventional wisdom that neurons communicate with each other based on single action potentials, represented as 1D functions (either binary or continuous).
Instead, it favors a more realistic view that neurons actually communicate by their firing patterns~\cite{rhythm_book,book_dynamical_systems_neuro}, such that a single neuron does not just communicate one single bit.
\cite{caml} employs probability as a universal language for describing the distributions of the firing patterns,  and uses embeddings (sufficient statistics) to represent their approximate forms.

One obvious advantage of this alternative view of deep neural networks is modeling power, as already demonstrated in~\cite{caml}.
Nevertheless, if we narrowly confine AI to defining compositions of activation functions, no matter how much meaning we give to them, they always fall into solving problems with very similar forms:
\begin{align}\label{equ:nn_loss}
\min_{\mathbf{\theta}=\{\theta_1, \ldots, \theta_n\}} \sum_{x\in\mathcal{D}} \mathcal{L}(x, \mathbf{\theta}) \equiv \sum_{x\in\mathcal{D}} f_1( f_2(\ldots f_n(x, \theta_n), \ldots; \theta_2), \ldots; \theta_1), n\in \mathbb{N},
\end{align}
where $\mathcal{D}$ denotes the whole or a mini-batch of training data.
The gradients of Eq.~\ref{equ:nn_loss} can be computed by applying the chain rule to the learnable parameter set $\theta_i$ for each $f_i, i = 1, \ldots, n$:
\begin{align}
\frac{\partial \mathcal{L}(x, \mathbf{\theta})} {\partial \theta_i} = \frac{\partial \mathcal{L}(x, \mathbf{\theta})} {\partial f_i} \frac{\partial f_i} {\partial \theta_i} = \frac{\partial \mathcal{L}(x, \mathbf{\theta})} {\partial f_1}\frac{\partial f_1} {\partial f_2} \ldots \frac{\partial f_{i - 1}} {\partial f_i} \frac{\partial f_i} {\partial \theta_i}
\end{align}
The algorithm for computing the gradients values of $\frac{\partial \mathcal{L}(x, \mathbf{\theta})} {\partial f_i}$ and $\frac{\partial \mathcal{L}(x, \mathbf{\theta})} {\partial \theta_i}$ recursively from $f_1$ down to $f_n$ is called \emph{back-propagation}.
Defining a loss function then solving it by the back-propagation algorithm is now a de facto approach in artificial neural networks.

From the above process, one can see that nothing prevents us from changing the dimensions of $x$ or $\theta_i, i\in \{1, 2, \ldots, n\}$ if the back-propagation algorithm is run on one batch at a time. 
However, the designs of existing deep learning libraries have not considered this as an essential feature. 
In the rest of this section, we propose a new framework that accounts for model growth.

\subsection{Need for growth}\label{subsec:need_for_growth}
A basic requirement for an intelligent system (biological or artificial) is the ability to process novel information from sensory input.
When handling a novel input in a neural network, it is necessary to convert it into a representation that can be processed by a loss function like Eq.~\ref{equ:nn_loss} with $x\in\mathbb{R}^m$.
In particular, if the input involves discrete objects such as words, it is \emph{necessary} to map them to an embedding space first.
A naive explanation for this necessity is from a neural network point of view: a discrete input $c$ can be represented as a characteristic (one-hot) vector $\vec{c}_{0/1} = [0, \ldots, 1, \ldots, 0]^T$, and through a linear activation layer it becomes $\mathbf{W} \vec{c}_{0/1} = \mathbf{W}_i$, where $\mathbf{W}_i$ represents the $i$th column of the real matrix $\mathbf{W}$, or equivalently, $c$'s embedding. Such an interpretation can explain what limits the implementations of deep neural networks with sparse input values and why a dictionary is always needed (\emph{i.e.}, a dictionary essentially defines $\mathbf{W}$).

In practice, the dimension of the characteristic vector $\vec{c}_{0/1}$ (\emph{i.e.}, number of columns in $\mathbf{W}$) can grow to be arbitrarily large and the embedding dimension (\emph{i.e.}, number of rows in $\mathbf{W}$) should also grow accordingly.
To see why the embedding dimension should grow, we resort to the sufficient statistics point of view for neural network layers~\cite{caml} and a basic fact that the value for each dimension of an embedding must be bounded\footnote{This is true for both machines, whose floating point representation is bounded by number of bits (\emph{e.g.}, 64), and biological systems, whose voltage or firing frequency of signals cannot exceed certain bound ($E=mc^2$).}.
This said, let us assume a layer of neural network represent the distribution $P(w|c)\propto\exp(\langle\vec{w}, \vec{c}\rangle)$.
Then two inputs $c_1$ and $c_2$ are considered different, if their corresponding distributions are sufficiently apart from each other.
Let $P_{c_1}(w) \equiv P(w| c_1)$ and $P_{c_2}(w) \equiv P(w | c_2)$, this can be represented as
\begin{align}\label{equ:embed_diversity}
D_{KL}(P_{c_1} || P_{c_2}) \equiv \int_w P(w | c_1) \frac{\log P(w | c_1)}{\log P(w | c_2)}> \delta,
\end{align}
where $D_{KL}(P|Q)$ denotes the KL-divergence between two distributions $P$ and $Q$, and $\delta > 0$ is a threshold.
By substituting the embedding form of $P(w|c)$, \emph{i.e.}, $P(w|c)\propto \exp(\langle \vec{w}, \vec{c}\rangle)$, into the above equation, we obtain:
\begin{align}
D_{KL}(P_{c_1} || P_{c_2})  \propto \int_w P(w|c_1) \langle\vec{w}, \vec{c}_1 - \vec{c}_2 \rangle.
\end{align}
Geometrically, it computes an average length of the vector $\vec{w}$ along the direction $\vec{c}_1 - \vec{c}_2$.
Since the length of $\vec{c}$ is bounded, the only way to always satisfy the inequality of Eq.~\ref{equ:embed_diversity} when the number of distinct $c$ increases is to increase the dimensions of $\vec{c}$ and $\vec{w}$.
Intuitively, it simply says that in order to fit more objects in a bounded space such that they are sufficient far apart from each other, one would have to expand to higher dimensions.

\subsection{A new neuron model: DynamicCell}
Now that we have addressed \emph{why} an AI system should grow, another imperative question is \emph{how}: how come a group of neurons, connected to each other only through input/output signals, work together to achieve an overall state of survival?
An ideal neuron model should not just explain how a single cell works, but also be generalizable to groups of cells (organisms), or even to groups of organisms. 
Even better, it should also explain the success of existing approaches widely used in deep learning, \emph{e.g.}, the back-propagation algorithm. 

\subsubsection{Motivations from free energy principle}

The free energy principle~\cite{free_energy_friston06} developed for understanding the inner workings of the brain provides us with some clues on how to build a more unified model for neural network learning.
In essence, it assumes a biological system is enclosed by a Markov blanket that separates the internal state from the external environment, and communications only occur through the sensory input and actions.
The overall goal of the biological system is to maintain a stable state (homeostasis), both internally and externally, or mathematically, to minimize the free energy (surprises) from inside and outside.

However, if an organism, enclosed by a Markov blanket, can only minimize the free energy through changing internal states and/or interacting with the environment, what if both means failed?
For example, when a person heard about a tragic news and there is no action can be taken, changing internal state can only disrupt the homeostasis of a body. 
Also from physics point of view, if information and energy are interchangeable~\cite{info_to_energy_nature} and the total energy is conserved, the discharge of excessive, non-digestive energy is also an essential way of maintaining homeostasis.

\begin{figure}[t]
\centering
\begin{tabular}{c}
    \mbox{\epsfig{figure= 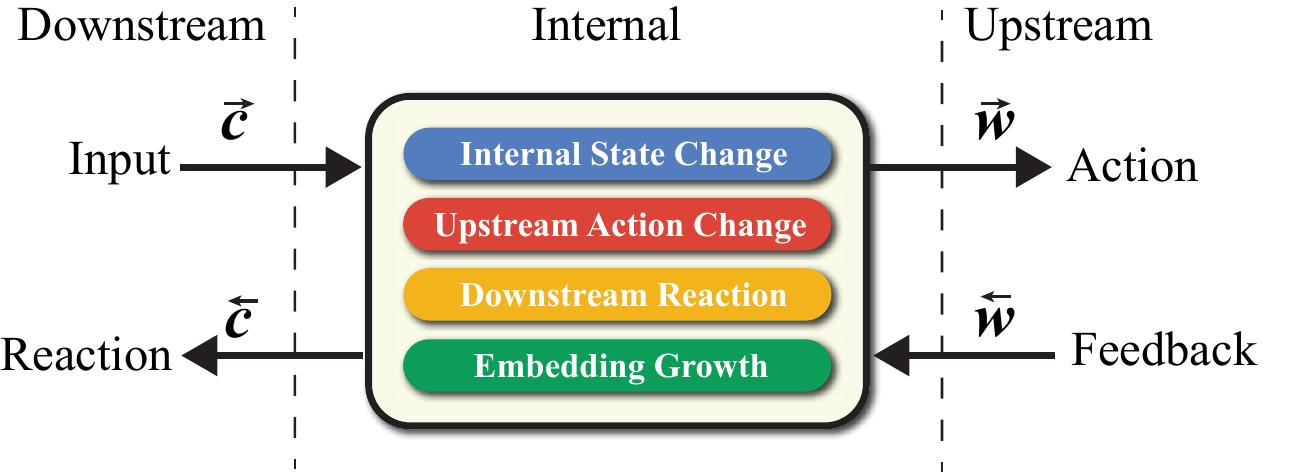, height = 3.2cm}}
\end{tabular}
\caption{\small In our DynamicCell model, we build a basic unit of life (cell) by introducing \emph{reaction} into the free energy principle. A basic activity of life is still to maintain its homeostasis. In addition, it should also ``react'' to unexpected input as a way to discharge the excessive energy that cannot be processed by changing internal states or actions. 
For example, laughing and crying are means to discharge good and bad surprises, respectively, which by themselves do not contribute to survival. In other words, \emph{life reacts}.} \label{fig:layer}
\end{figure}

Hence, we can simply improve the idea of free energy principle by including \emph{reaction} into the picture (Fig.~\ref{fig:layer}), which is essential according to the law of conservation of energy in physics.
In our new model, each cell or a group of them (organism) can act according to the same principle: minimize the free energy (surprise from input $\vec{c}$) through change of internal states and/or actions, and the excessive non-digestive energy that cannot be minimized is discharged via reaction.
Here the action signal $\vec{w}$ is received by other upstream cells inside the same Markov blanket, and can only affect upstream feedback $\overset{{}_{\shortleftarrow}}{w}$.
Note that the action signal $\vec{w}$ here is different from physical actions taken by an organism to interact with the environment.
Under our model, physical actions can be activated either by upstream signal $\vec{w}$ (\emph{e.g.}, to motor neurons) to do useful work (\emph{e.g.}, running away from threats), or by downstream signal $\overset{{}_{\shortleftarrow}}{c}$ to discharge extra surprises (\emph{e.g.}, via laughing or crying).

\subsubsection{Formulation}

To formulate the above ideas in a mathematical form, we resort to~\cite{caml} again as a starting point for building a new neuron model.
Overall, a neuron represents the distribution $P(w|c)$ and according to~\cite{caml}, its input and output signals can be approximately represented by their embeddings, \emph{e.g.},
$P(w|c) = \frac{1}{Z(\vec{c})}\exp(\langle\vec{w}, \vec{c}\rangle)$ where $\vec{w}$ may depend on $\vec{c}$, and $Z(\vec{c}) = \sum_{\vec{w}} \exp(\langle\vec{w}, \vec{c}\rangle)$.
Given this assumption, we can represent the minimization of free energy, or surprise, into two parts: internal and external.

\paragraph{Internal state homeostasis}
The stability of a cell's internal state is reflected in the action state $\vec{w}$ in Fig.~\ref{fig:layer}.
The long-term behavior of a cell should be independent of its context $c$ (context-free) and therefore can be represented as the distribution $P_{\vec{w}}\equiv P(w)$.
Hence the free energy, or surprise, on the internal state of a cell from a given input $c$ can be simply represented as
\begin{align}\label{equ:kl_surprise}
D_{KL}(P_{\vec{w}} || P_{c}) = \sum_x P_{\vec{w}} (x) \log \frac{P_{\vec{w}} (x)}{P(x|c)},
\end{align}
and surprise minimization means adjusting internal parameters of $P(w|c)$ such that $P(w|c) \approx P(w)$.
To see how surprise minimization can be implemented in the embedding space, let us apply the sufficient statistics representation of $P(w|c)$ and rearrange Eq.~\ref{equ:kl_surprise} as follows:
\begin{align}\label{equ:kl_surprise_exp}
D_{KL}(P_{\vec{w}} || P_{c}) = - \sum_x P_{\vec{w}}(x) \langle\vec{w}, \vec{c}\rangle + \log Z(\vec{c}) - H(P_{\vec{w}})
\end{align}
where $H(\cdot)$ denotes the entropy of the given distribution and should be relatively stable.
To minimize Eq.~\ref{equ:kl_surprise_exp}, a cell needs to reach a state where the gradient of $D_{KL}(P_{\vec{w}} || P_{c})$ with respect to input $c$ is zero:
\begin{align}\label{equ:kl_derivative}
\frac{\partial D_{KL}(P_{\vec{w}} || P_{c})}{\partial \vec{c}} = 0 
&\Leftrightarrow
 - \sum_x P_{\vec{w}}(x) \frac{\partial \langle\vec{w}, \vec{c}\rangle}{\partial \vec{c}} + \frac{\partial \log Z(\vec{c})}{\partial \vec{c}}  \approx 0 \nonumber \\
 &\Leftrightarrow
  \langle\vec{w}\rangle_{P_{c}}\approx \langle\vec{w}\rangle_{P_{\vec{w}}} 
\end{align}
where we assume $\partial \langle\vec{w}, \vec{c}\rangle / \partial \vec{c} \approx \vec{w}$.
Note that this is similar in form as the contrastive divergence algorithm~\cite{contrast_divergence_hinton}, though they are derived based on completely different assumptions (\cite{contrast_divergence_hinton} assumes a neuron outputs 0/1 action potentials whereas in our case a neuron outputs the embedding that represents the distribution of its firing patterns).

\paragraph{Upstream state homeostasis}
The difference between upstream and downstream is that the state of the former is expected to be stable\footnote{From the point of view of behavioral psychology, this is consistent with an organism's natural tendency to distinguish ``us'' from ``them''~\cite{behave_book}.}.
To measure the stability of upstream states, one can treat the whole complex process of information processing in the upstream as a blackbox and simply measure its deviation from usual distribution, namely:
\begin{align}\label{equ:kl_surprise2}
D_{KL}(P_{\overset{{}_{\shortleftarrow}}{w}} || P_{\vec{w}}) = \sum_x P_{\overset{{}_{\shortleftarrow}}{w}}(x) \log \frac{P_{\overset{{}_{\shortleftarrow}}{w}}(x)}{P(x|w)},
\end{align}
where $P_{\overset{{}_{\shortleftarrow}}{w}}$ represents the distribution of the upstream feedback signal $\overset{{}_{\shortleftarrow}}{w}$ (Fig.~\ref{fig:layer}).
Similar to Eq.~\ref{equ:kl_derivative}, we can obtain the condition for stable upstream state as:
\begin{align}\label{equ:kl_derivative2}
\frac{\partial D_{KL}(P_{\overset{{}_{\shortleftarrow}}{w}} || P_{\vec{w}})}{\partial \vec{w}} = 0 \Leftrightarrow 
\langle \overset{{}_{\shortleftarrow}}{w} \rangle_{P_{\vec{w}}} \approx \langle\overset{{}_{\shortleftarrow}}{w}\rangle_{P_{\overset{{}_{\shortleftarrow}}{w}}}.
\end{align}
Hence by changing internal state of $P(w|c)$, a cell can optimize both Eq.~\ref{equ:kl_surprise_exp} and Eq.~\ref{equ:kl_surprise2} to minimize its overall surprise. 
An equilibrium is a balance between internal state and actions.

\paragraph{Reaction}
From the above analysis, the free energy is minimized when both Eq.~\ref{equ:kl_derivative} and~\ref{equ:kl_derivative2} are satisfied.
However, it is a natural tendency for the entropy of the overall state of a system to increase, so an enclosed organic system should expect constant upcoming surprises from the input.
When these surprises cannot be minimized by changing internal states (Eq.~\ref{equ:kl_derivative}) or taking actions (Eq.~\ref{equ:kl_derivative2}), they must be discharged out of the system in some way, \emph{i.e.}, through reaction $\overset{{}_{\shortleftarrow}}{c}$.
For example, one choice of the total additional energy can be represented as 
\begin{align}\label{equ:reaction}
\overset{{}_{\shortleftarrow}}{c} \approx  
(|\langle\overset{{}_{\shortleftarrow}}{w}\rangle_{P_{\overset{{}_{\shortleftarrow}}{w}}} - \langle \overset{{}_{\shortleftarrow}}{w} \rangle_{P_{\vec{w}}}|^2
+ |\langle\vec{w}\rangle_{P_{\vec{w}}} - \langle\vec{w}\rangle_{P_{c}}|^2)/2
\geq
(\langle\overset{{}_{\shortleftarrow}}{w}\rangle_{P_{\overset{{}_{\shortleftarrow}}{w}}} - \langle \overset{{}_{\shortleftarrow}}{w} \rangle_{P_{\vec{w}}})
\circ (\langle\vec{w}\rangle_{P_{\vec{w}}} - \langle\vec{w}\rangle_{P_{c}}),
\end{align}
where $|\cdot|^2$ represents the element-wise square and $\circ$ is also an element-wise product.
In the following, we will see that this form is chosen purely for the convenience of connecting it to gradient descent update for loss functions.
There exists many other possibilities in defining reaction, which is not a major focus of this paper.

\paragraph{Connection to gradient descent update}
Finally, let us take a look at how the above process subsumes conventional loss minimization using gradient descent as its special case.
To see this, we can simply wire the action signal $\vec{w}$ to a loss function $\mathcal{L}(\vec{w})$ and let $\overset{{}_{\shortleftarrow}}{w}$ return the evaluation of the loss (\emph{i.e.}, $\overset{{}_{\shortleftarrow}}{w} = \mathcal{L}(\vec{w})$).
From the above relations and by taking the finite difference step to be $1$ in gradient approximation, we obtain
\begin{align}
\frac{\partial D_{KL}(P_{\vec{w}} || P_{c})}{\partial \vec{c}} &\approx \langle\vec{w}\rangle_{P_{\vec{w}}} - \langle\vec{w}\rangle_{P_{c}} \approx \frac{\partial \vec{w}}{\partial \vec{c}} \\
\frac{\partial D_{KL}(P_{\overset{{}_{\shortleftarrow}}{w}} || P_{\vec{w}})}{\partial \vec{w}} &\approx 
\langle\overset{{}_{\shortleftarrow}}{w}\rangle_{P_{\overset{{}_{\shortleftarrow}}{w}}} - \langle \overset{{}_{\shortleftarrow}}{w} \rangle_{P_{\vec{w}}}
\approx
\langle \mathcal{L}(\vec{w})\rangle_{P_{\overset{{}_{\shortleftarrow}}{w}}} - \langle \mathcal{L}(\vec{w}) \rangle_{P_{\vec{w}}}
\approx \frac{\partial \mathcal{L}(\vec{w})}{\partial \vec{w}}
\end{align}
Finally, from Eq.~\ref{equ:reaction}, we arrive at the familiar form of gradient:
\begin{align}
\overset{{}_{\shortleftarrow}}{c} \approx   \frac{\partial \mathcal{L}(\vec{w})}{\partial \vec{w}} \cdot \frac{\partial \vec{w}}{\partial \vec{c}}   = \frac{\partial \mathcal{L}}{\partial \vec{c}}
\end{align}
This is consistent with progresses in cognitive science that real brains actually do some form of back-propagations~\cite{back_propagation_in_brain}.

\section{System Design}

\subsection{Tensorflow API Design}
\label{section:tensorflow_api}
Recall that now each layer/neuron in a neural network is considered to represent certain distribution $P(w|c)$ in the embedding space with $c$ as the input and $w$ the output.
For intermediate layers between input and output, $c$ is already represented as an embedding $\vec{c}$ and one only needs to define a function that computes 
$\vec{w}$. 
In such a case, we can just use the same computational graph in TensorFlow for forward (Input and Action in Fig.~\ref{fig:layer}) and backward (Feedback and Reaction in Fig.~\ref{fig:layer}) executions, and non-gradients based update can be achieved via slight changes to the function \emph{tf.gradients}. 
For example, a typical DynamicCell node can be defined as
\begin{lstlisting}[language=Python]
def my_cell_forward(c):
  """Returns action w"""
  
@ops.RegisterGradient("MyCellForward")
def my_cell_backward(op, w_feedback):
  """Returns reaction c_feedback"""
\end{lstlisting}

However, special cares are needed when one of $w$ and $c$ involves sparse features (\emph{e.g.}, words), as it may happen at the input or output layer (\emph{e.g.}, a softmax output layer that predicts a word).
Existing TensorFlow implementation always requires a dictionary and string-to-index conversions (\emph{e.g.}, via \emph{tf.nn.embedding\_lookup} or \emph{tf.math.top\_k}), which is incompatible with our philosophy that the users only need to define the form of $P(w|c)$ without worrying about its content.
In fact, these input/output operations are the key to enabling Tensorflow to handle ever-growing distinct input/output values by transferring the job of content processing to DynamicEmbedding service (DES).
In addition, to let Tensorflow seamlessly work with DES, we use a single protocol buffer to encode all the configurations, which is represented as the input argument \emph{de\_config} in our Tensorflow APIs.

\subsubsection{Sparse input}\label{subsec:sparse_input}
As mentioned above, allowing TensorFlow to directly take any string as input can be very beneficial.
Here we call the process to convert any string input into its embedding \emph{dynamic embedding}, with its TensorFlow API defined as
\begin{lstlisting}[language=Python]
def dynamic_embedding_lookup(keys, de_config, name):
  """Returns the embeddings of given keys."""
\end{lstlisting}
where \emph{keys} is a string tensor of any shape, and \emph{de\_config} contains the necessary information about the embedding, including the desired embedding dimension,
initialization method (when the key is first seen), and storage of the embedding, etc. 
Also the \emph{name} and the config can uniquely identify the embedding to facilitate data sharing.

\subsubsection {Sparse output}\label{subsec:sparse_output}
When a neural network outputs sparse features, it is usually used in an inference problem: $\argmax_{w} P(w|c)$, where $c$ is input from previous layer, represented as $\vec{c}$ in a neural network. 
According to Sec.~\ref{subsec:exp_family}, if we assume $P(w|c) \propto \exp(\langle\vec{w}, \vec{c}\rangle)$ where $\vec{w}$ is the embedding of $w$, then $\argmax_{w} P(w|c) = \argmax_{w} \langle\vec{w}, \vec{c}\rangle$, which is simply the closest point (measured by the dot product) to the input query $\vec{c}$ among the embeddings of all the values of $w$.
In fact, the softmax function that is commonly used in neural network is closely related to our formulation. 
To see this, let us assume the set of all possible values of $w$ to be $\mathcal{W}$, and $\forall a\in\mathcal{W}$, the softmax probability can be represented as
\begin{align}\label{equ:softmax}
P(w = a | c) = \frac{\exp(\vec{c}^T \vec{w}_a + b_a)}{\sum_{k\in\mathcal{W}}\exp(\vec{c}^T \vec{w_k} + b_k)}
= \frac{\exp(\langle  \begin{bmatrix}\vec{c} \\1\end{bmatrix}, \begin{bmatrix}\vec{w}_a \\ b_a\end{bmatrix}\rangle)}{\sum_{k\in\mathcal{W}}\exp(\langle \begin{bmatrix}\vec{c} \\ 1\end{bmatrix}, \begin{bmatrix}\vec{w}_k\\ b_k\end{bmatrix}\rangle)},
\end{align}
which falls into our special case if we let $dim(\vec{w}) = dim(\vec{c}) + 1$ where $dim(\cdot)$ denotes the dimension of a vector.

However, when the number of elements in $\mathcal{W}$ is very large, it is inefficient to compute the cross entropy loss for softmax output as it requires the computation of Eq.~\ref{equ:softmax} for all values of $w$.
Fortunately, efficient negative sampling method has already been well studied~\cite{softmax_sampling}. All we have to do is to support it in DES.

\paragraph{Candidate negative sampling} To allow potentially unlimited number of output values, we follow the implementation of \emph{tf.nn.sampled\_softmax\_loss} to define an internal function that returns the logit values ($\vec{c}^T \vec{w_k} + b_k, w_k\in \mathcal{W}_{sampled}$) of negative samples from given activation positive keys and $\vec{c}$.
The API can be defined as 
\begin{lstlisting}[language=Python]
def _compute_sampled_logits(pos_keys, c, num_sampled, de_config, name):
  """Returns sampled logits and keys from given positive labels."""
\end{lstlisting}
Here \emph{num\_sampled} is a positive number and the sampling strategy is defined in \emph{de\_config}.

\paragraph{TopK retrieval} Whereas candidate negative sampling is needed during training, during inference, we would like to compute $\argmax_{w}  P(w|c) = \argmax_{w} \langle\vec{w}, \vec{c}\rangle$ as mentioned above, and in practice it is more common to retrieve the top-k nearest points to given input (\emph{e.g.}, for beam search in language inference). The interface for TopK retrieval is defined as
\begin{lstlisting}[language=Python]
def top_k(c, k, de_config, name):
  """Returns top k closest labels to given activation c."""
\end{lstlisting}
Behind the scene, the function should call the DynamicEmbedding server to find the keys whose embeddings are closest to $\begin{bmatrix}\vec{c} \\1\end{bmatrix}$. 

\subsubsection {Saving/restoring models}
Finally, during model training, a model needs to be periodically saved.
Since we have moved most of model data out of TensorFlow's graph, it is important to maintain the consistency between the checkpoints saved by both TensorFlow and DynamicEmbedding.
On the API side, each time a DynamicEmbedding related API is called, the corresponding embedding data information, uniquely identifiable by (name, \emph{de\_config}) should be stored in a global
variable.
Checkpoint saving/loading for DynamicEmbedding would be very similar to TensorFlow, namely, 
\begin{lstlisting}[language=Python]
save_path = save(path, ckpt)
restore(save_path)
\end{lstlisting}

If the user uses the automatic training framework in TensorFlow, such as tf.estimator, saving/loading the model is automatically handled by our high-level APIs. But if they want to do it in a low level fashion, they will need to call the above functions along with TensorFlow's corresponding I/O functions.

\subsubsection {A Word2Vec model with DynamicEmbedding}
In summary, the TensorFlow API change to support DynamicEmbedding is very straightforward, and the job for model construction is also simplified.
As an example, a Word2Vec model can be defined with only a few lines of code:
\begin{lstlisting}[language=Python]
tokens = tf.placeholder(tf.string, [None, 1])
labels = tf.placeholder(tf.string, [None, 1])
emb_in = dynamic_embedding_lookup(tokens, de_config, 'emb_in')
logits, labels = _compute_sampled_logits(labels, emb_in, 10000,  de_config, 'emb_out')
cross_ent = tf.nn.softmax_cross_entropy_with_logits_v2(labels, logits)
loss = tf.reduce_sum(cross_ent)
\end{lstlisting}
Note that the need for a dictionary is completely removed.

\subsection{DynamicEmbedding Service Design}
\label{section::des}
As shown in Fig.~\ref{fig:overview}, our DynamicEmbedding Service (DES) involves two parts: DynamicEmbedding Master (DEM) and DynamicEmbedding Workers (DEWs).
The TensorFlow APIs defined in the previous section only communicate with DEM, which in turn distributes the real work into different DEWs.
To achieve both efficiency and ever growing model, each worker in DEWs should balance between local caching and remote storage.
In this section, we discuss different aspects of DES in its current form.

\subsubsection{Embedding storage}\label{subsec:embedding_store}
As discussed in Sec.~\ref{section:math_formulation}, communications among neurons are represented as sufficient statistics of their firing patterns (embeddings), which are simply vectors of floating values. These firing patterns are discrete in nature and can be represented by string ids.
Hence the storage of these embedding data only involves (key, value) pairs, and not surprisingly, we use protocol buffer to facilitate data transfer and keep additional information for each embedding like string id, frequency, etc.

When certain data is passed into a node defined by our TensorFlow API, it would talk to DES to do the actual job.
For example, during the forward pass of running the \emph{dynamic\_embedding\_look} op (Sec.~\ref{subsec:sparse_input}), a batch of strings are passed into a node in the computation graph of TensorFlow, and it in turn asks DEM to process the actual lookup job.
During the backward pass, the feedback signal (\emph{e.g.}, gradients with respect to its output) is passed into the registered backward node and it also needs to talk to DEM for data update. 
 
To allow for scalable embedding lookup/update, we design a component called EmbeddingStore that is dedicated to communicating with multiple storage systems available inside Google. 
Each supported storage system implements the same interface with basic operations like Lookup(), Update(), Sample(), Import(), Export().
For example, an \emph{InProtoEmbedding} implements the EmbeddingStore interface by saving the whole data into a protocol buffer format, which can be used for local tests and training small data set.
An \emph{SSTableEmbedding} loads the data into the memories of DEWs during training and saves them to immutable but potentially very large files in Google File System (GFS)~\cite{gfs}.
A  \emph{BigtableEmbedding} allows the data to be stored both in local cache and remote, mutable and highly scalable Bigtables~\cite{bigtable}. therefore enabling fast recovery from worker failure as it does not need to wait until all the previous data are loaded before accepting new requests.

\subsubsection{Embedding update}
In our framework, embedding updates may happen during both forward and backward passes (Fig.~\ref{fig:layer}).
For the backpropagation algorithm, updates only occur when backward feedback with the information $\partial \mathcal{L} / \partial w$ arrives.
To guarantee that our system is completely compatible with any existing gradient descent algorithms (\emph{e.g.}, \emph{tf.train.GradientDescentOptimizer} or \emph{tf.train.AdagradOptimizer}), we need to implement each of them inside DEWs.
Fortunately, we can simply reuse the same code implemented in TensorFlow to guarantee consistency.
One caveat is that many gradient descent algorithms, such as Adagrad~\cite{tf_adagrad}, keeps global information about each value that should be consistent between gradient updates.
In our case, this means we need additional information stored in the embedding.

\paragraph{Long-short term memory}
When a learning system is capable of processing data that spans a long period (\emph{e.g.}, months or years), it is important to address the topic of long-short term memory since if certain features only show up momentarily or have not been updated for a long time, it may not help with inference accuracy.
On the other hand, some momentary input may contains valuable information that needs special treatment, an unbiased learning system should be able to process those low frequency data.
In the following, we propose two basic techniques for managing the lifetime of the embedding data.
\begin{itemize}
\item \emph{Frequency cutoff}: Each time an embedding is updated, a counter is incremented to record its update frequency.
Therefore we can decide if this embedding should be saved to a permanent storage (\emph{e.g.} Bigtable) based on a cutoff value on its frequency.
For training that involves multiple epoches, it is TensorFlow's job to tell if an example is seen for the first time.

\item \emph{Bloom filter}: Another popular approach that achieves similar effect of pruning off low frequency data, only more memory efficient, is to use bloom filter~\cite{bloom_filter}. We implemented this feature also for the purpose of compatibility with existing linear systems that can already handle large amount of data, but with much less complex models than deep networks.
\end{itemize}


\subsubsection{Top-k sampling}
During model inference, it is important to efficiently retrieve the top k closest embeddings from given input activation, where the distance is measured by dot product as shown in Sec.~\ref{subsec:sparse_output}.
This can be efficiently done both accurately and approximately for very large number of input (\emph{e.g.} \cite{scam_nips17}). 
We employ existing implementations available inside Google to let each worker in DEWs returns its own top k embeddings to DEM.
Assuming there are $n$ DEWs, then DEM would select the top-k closest points among the $n\times k$ candidate vectors.
This way, both accuracy and efficiency are guaranteed if $k \ll m$ where $m$ is the total number of keys.

\subsubsection{Candidate sampling}
Sampling can be tricky if they are stored in remote storage such as Bigtable. 
This is why we need metadata to store the necessary information for us to efficiently sample candidates. 
At the very least we should support sampling strategies used by two existing Tensorflow ops: \emph{tf.nn.sampled\_softmax\_loss} (based on a partition strategy) and \emph{tf.contrib.text.skip\_gram\_sample} (based on frequency). 
If we want to achieve even better word embedding, higher order information such as PMI (probability of mutual information) or count of cooccurrence should also be computed or sampled accordingly (\cite{emb_glove}). 
Therefore, these bookkeeping information needs to be processed during embedding lookup for efficient sampling.

Here we decide to re-implement candidate sampling in DES due to the following concerns:
(i) It is not easy to reuse TensorFlow code as they assume every embedding has a unique index in an integer array.
(ii) The original implementation does not consider multiple label output, due to the fact it separates true labels and sampled labels (To meet the constraint that all variables must be defined before training, it requires number of true labels from the input such that each input must have exactly the same true labels. This is an overly strict requirement for many realistic, multiple-label training).

In our new design, to still meet the requirement that the graph should be fixed while allowing varying number of true\_labels from each input, we simply merge positive and negative examples and let the user decide the value of num\_samples. Our interface becomes:
\begin{lstlisting}[language=C]
class CandidateSampler {
 public:
  struct SampledResult {
    string id;
    bool is_positive;
    float prob;
  };
  std::vector<SampledResult> Sample(
      const std::vector<string>& positive_ids, const int num_sampled,
      const int range) const;
};
\end{lstlisting}
Therefore, our new candidate sampling also addresses a limitation in TensorFlow's implementation, leading to better handling of multi-label learning.

\subsubsection{Distributed computation}
Distribution is straightforward given that each embedding data requires a unique string id as its lookup key (Sec.~\ref{subsec:embedding_store}).
Each time the DEM receives a request from the TensorFlow API it partitions the data based on their ids and distribute the work into different DEWs (lookup, update, sample, etc).
Here each worker is responsible for a unique subset of the total universe of keys, and after recovered from machine failures, it should still be able to identify the subset of keys it is responsible for.
This is possible with Google's Borg system~\cite{borg} as each worker inside a server can uniquely identify its own shard number.
For example, when there are $n$ workers, the $i$th worker ($i = 0, 1, \ldots, n - 1$) would be responsible for those embeddings with keys satisfying $mod(hash(key), n) = i$.
For efficient candidate sampling, DEM should bookkeep metadata about each worker and decide the number of samples needed from each worker.

\subsubsection{Serving with scale}
Serving a TensorFlow model with DynamicEmbedding needs some special care since the embedding data need to be retrieved efficiently for large size (>$100$G). It is no longer feasible to fit them in local machines. 
Therefore, besides the most straightforward serving by DynamicEmbedding service, we should also support more robust design to handle large embedding data. 
Towards robust model serving, the following two optimizations are considered.

\paragraph{Sandbox mode}
During model serving, all the embedding data sources are known and the external storage is not updated. 
Therefore, it is reasonable that we provide some protection to prevent the server from serving other requests and accidentally update the storage.
Also our training is usually done stage by stage to guarantee data quality by version control.
The solution is to provide a sandbox mode for DynamicEmbedding server, using a few flags to control the source of data.
This way, no other training job is able to call the DynamicEmbedding server.

\paragraph{Remote storage lookup with local cache}
If the training job has already saved the embedding data into a remote storage system, this scheme allows serving without starting any servers from DynamicEmbedding, \emph{i.e.}, it becomes a standalone TensorFlow model.
Specifically, a local process can play the role of DEM and each embedding lookup directly talks to remote storage system.
Also the embedding data is stored in local cache, whose size depends on the available memory for TensorFlow jobs. 
To further reduce latency, it is important that we group a batch of keys into one single lookup call. 

\section{Experimental Results}
\label{section:results}
Our DynamicEmbedding system is evaluated on a few aspects: (i) \emph{Backward compatibility} with TensorFlow for gradient descent based models (Sec.~\ref{subsec:backward_compatibility}); (ii) \emph{Model accuracy} compared with models with limited dictionary size (Ses.~\ref{subsec:model_accuracy}); and (iii) \emph{System performance} in terms of memory usage and training speed (Sec.~\ref{subsec:system_performance}). In the end, we will present a colossal-scale model for keyword targeting in Sec.~\ref{subsec:megabrain}.

\subsection{Backward compatibility with TensorFlow for gradient updates}\label{subsec:backward_compatibility}
First of all, we would like to verify the correctness of our implementation by showing that our system is completely backward compatible with TensorFlow among gradient descent based tasks.
Recall that DynamicEmbedding not only takes care of the embedding lookup of each layer, but is also responsible for the embedding update, which includes gradient descent as its special case.
From a library user's perspective, this should not bring any difference between TensorFlow with and without DynamicEmbedding, in terms of training results.

We verify that this is indeed the case for the Word2Vec model~\cite{emb_mikolov2013word2vec} with three different gradient descent algorithms: SGD~\cite{sgd}, Adagrad~\cite{tf_adagrad}, and Momentum~\cite{gradient_descent_momentum}.
To guarantee that both TensorFlow and DynamicEmbedding run exactly the same model, we let DynamicEmbedding Manager take the same dictionary used by the TF-Word2Vec as its input filter and only process those input from the given dictionary. 
For all other input, they would be mapped to a special input, \emph{i.e.}, ``oov'' (out-of-vocab) -- using the same trick as existing dictionary based models.
We also guarantee that all other hyper-parameters, such as learning rate, batch size, are exactly the same between them.

\begin{figure}[t]
\centering
\begin{tabular}{ccc}
    \mbox{\epsfig{figure= 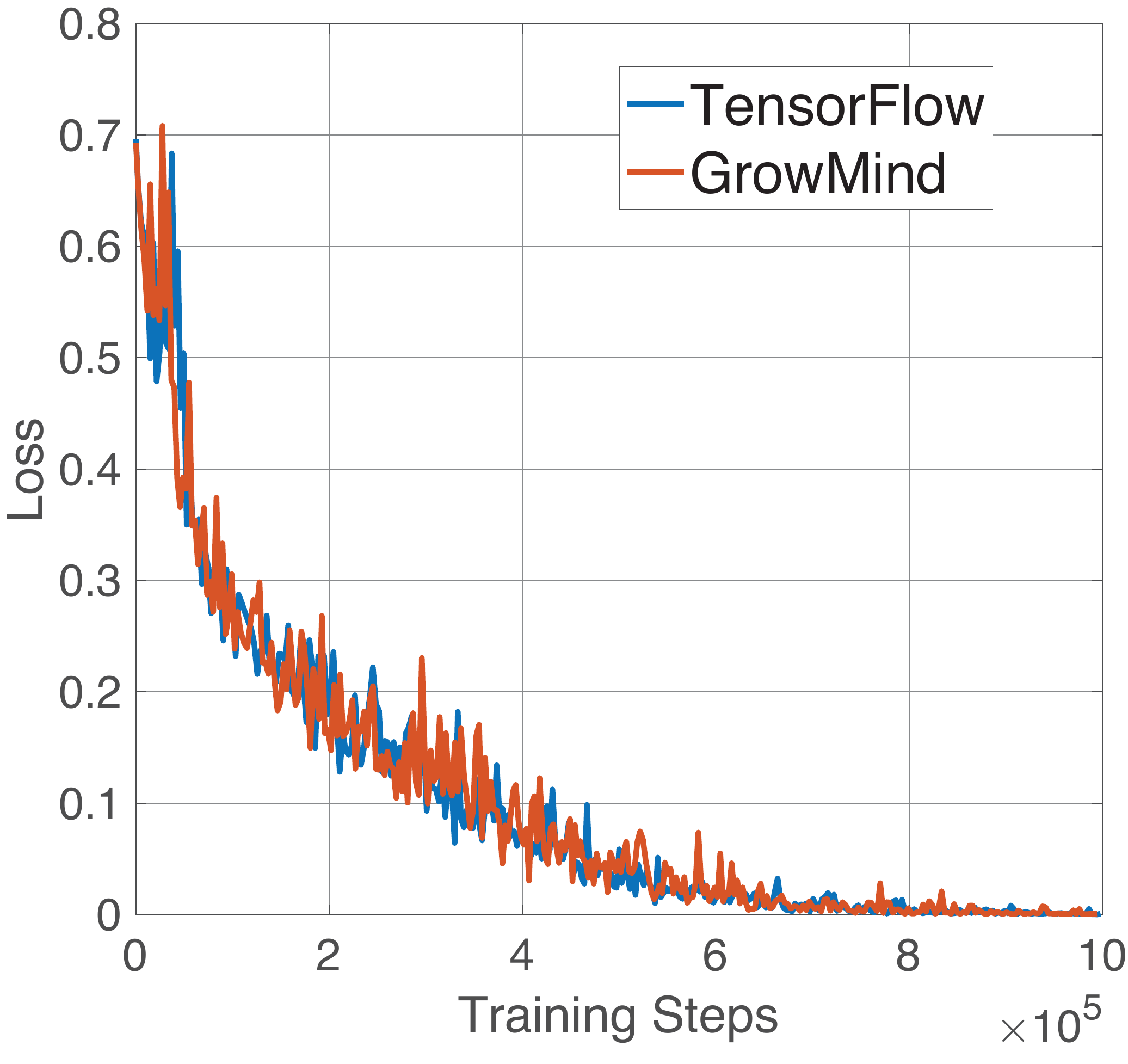, height = 4cm}} &
    \mbox{\epsfig{figure= 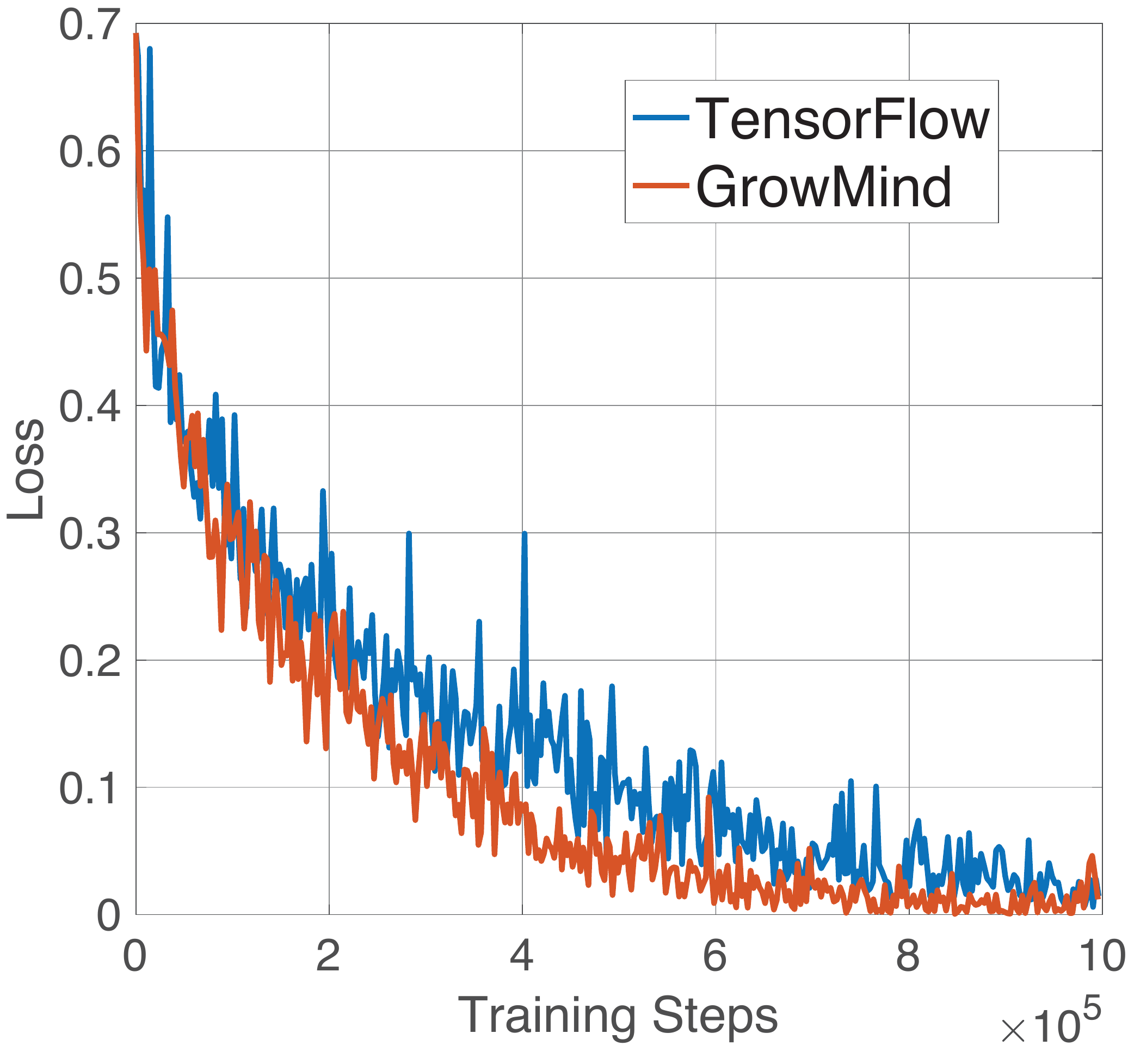, height = 4cm}} &
    \mbox{\epsfig{figure= 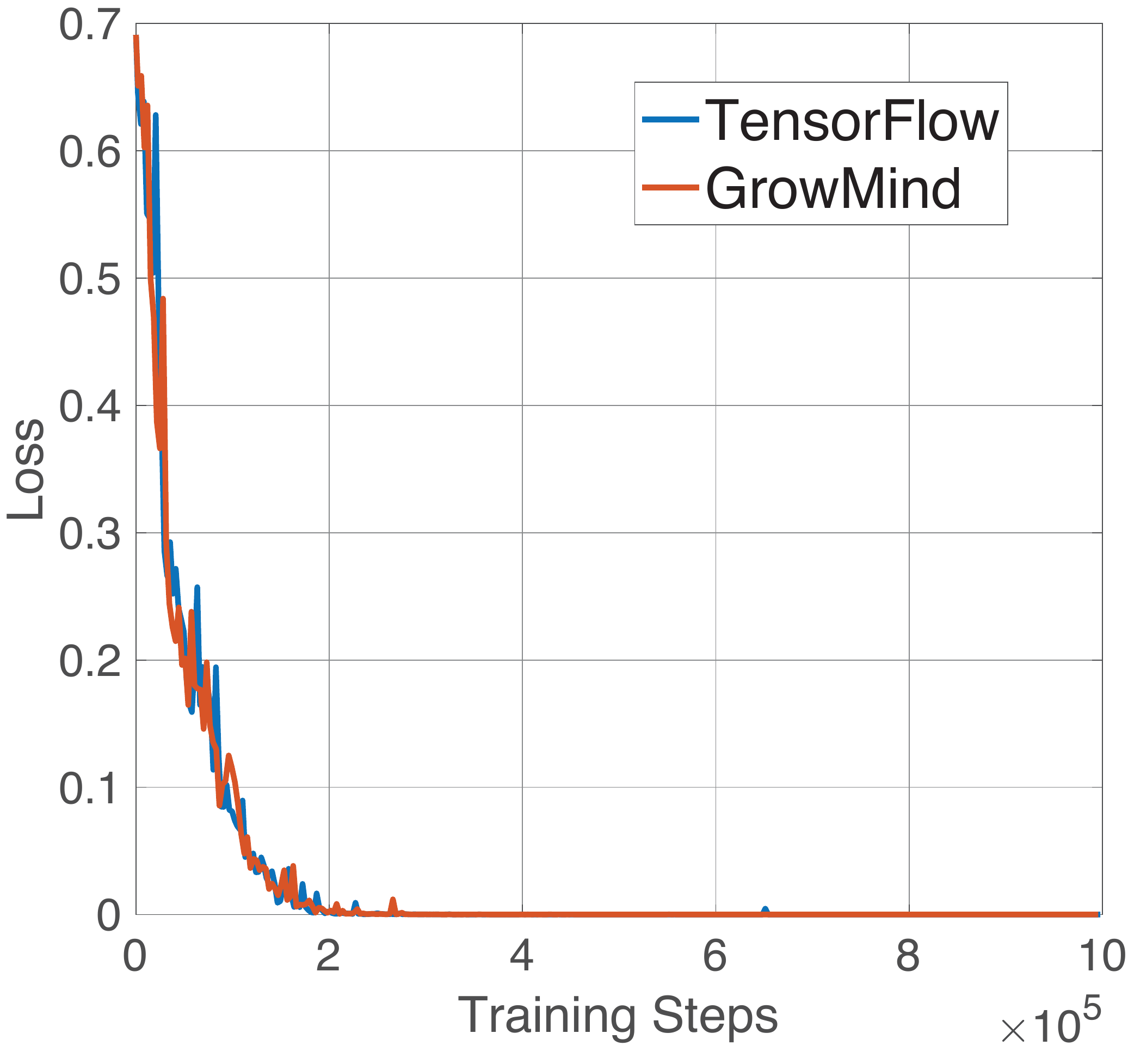, height = 4cm}} \\
    {\scriptsize (a) SGD~\cite{sgd}} & {\scriptsize (b) AdaGrad~\cite{tf_adagrad}} & {\scriptsize (c) Momentum~\cite{gradient_descent_momentum}} 
\end{tabular}
\caption{\small (Best viewed in color) Comparison between TensorFlow and DynamicEmbedding in solving the same Word2Vec model on the same training dataset using different gradient descent algorithms, which verifies DynamicEmbedding is backward compatible.} \label{fig:gradient_descent}
\end{figure}

Fig.~\ref{fig:gradient_descent} illustrates the comparisons in accuracy among three different gradient descent algorithms.
Because the embeddings are initialized randomly, we cannot expect their accuracy be matched perfectly. 
Nevertheless, for the first 1M training steps, both TensorFlow and DynamicEmbedding yield very close results for all three gradient descent algorithms.

\subsection{Model benchmarks}\label{subsec:model_accuracy}

Ever since DynamicEmbedding was first developed around early 2018, we have upgraded a number of popular models to enjoy ever growing feature size.
The following is a partial list of models we have built.

\paragraph{DE-Word2Vec} The skip-gram model~\cite{emb_mikolov2013word2vec} is upgraded by replacing its input with our dynamic embedding layer (Sec.~\ref{subsec:sparse_input}) and its output with our sampled logits layer (Sec.~\ref{subsec:sparse_output}). It is therefore capable of mapping any word (n-grams), regardless of language, into the same high-dimensional vector space. This model can map any search query into an embedding to upgrade Google Search's related search\footnote{\url{https://support.google.com/trends/answer/4355000?hl=en}} feature based on semantic rather than syntactic similarity.

\paragraph{DE-Seq2Seq} A seq2seq model predicts a sentence from another input sentence (\emph{e.g.}, in translation). Here we have upgraded the Google Neural Machine Translation (GNMT) model~\cite{gnmt} with attention mechanism to take a sequence of words as input (implemented by our dynamic embedding layer) and output a sequence of words (the output layer is implemented by our sampled logits layer). This model allows us to improve Google Translate by including a much larger vocabulary than previously used.

\paragraph{DE-Sparse2Seq} This is in essence similar to the Google Transformer model~\cite{attention_all_you_need}, which considers input as a bag of words and the output is generated by a recurrent neural network with attention mechanism. Similar to the two models mention above, we replace both input and output of the original model with our corresponding new DynamicEmbedding components. We applied this model to suggest ad headlines and descriptions\footnote{\url{https://support.google.com/google-ads/answer/1704389?hl=en}} from sparse features on a website and it can cover all languages using a single model.

\paragraph{DE-Image2Label} The input to the model is an image and the output is a label (\emph{e.g.} a phrase or a description). In our new model, we first let each image go through the Inception model~\cite{inception} and connect its output (an embedding of the image) to our sampled logits layer (Sec.~\ref{subsec:sparse_output}). This way, our model can be trained on arbitrary (image, label) pairs. We have applied this model to suggest a very large set of image labels available from Google Image Search, whose distinct label set is in the order of hundreds of millions. 

\paragraph{DE-Sparse2Label} An image is a highly structured input as it can always be normalized into a fixed size. However, there is a plethora of unstructured data in the world that contains multiple possible input sources. For example, a webpage may contain texts, tables, images organized in a hierarchical way and there are many potential labeling for its content.
Our DE-Sparse2Label model maps each of the source labels (may contain an unknown number of sparse values) into an embedding using the CA-SEM model~\cite{caml} and let it go through a few ResNet-like layers (\emph{i.e.}, CA-RES model~\cite{caml}) for further processing, then output a sparse label (\emph{e.g.}, a query that leads to this webpage). Here both the input and output use our DynamicEmbedding components to achieve ever-growing sparse feature size. This model has been successfully applied in our ad keyword suggestion system and its detail is discussed in Sec.~\ref{subsec:megabrain}.

\paragraph{DE-BERT} This is the DynamicEmbedding counterpart of the BERT model~\cite{bert} for natural language processing (NLP), which no longer requires a predefined maximal input length (\emph{e.g.}, $512$ or $1024$) or preprocessing the input into wordpieces~\cite{gnmt}. Our model is able to learn from any continuously growing text corpus (\emph{e.g.} online articles) and accordingly an ever-growing embedding repository is maintained for various downstream NLP applications.

\subsubsection{Model accuracy}
Now let us turn to another important aspect of DynamicEmbedding: model accuracy, and address the following question: Can a model with sufficiently large capacity bring better accuracy?

Theoretically, the answer is affirmative as more data simply means more experience and therefore a major concern in machine learning, \emph{i.e.} generalization, would be eliminated if the training data can already cover most cases. 
However, this is based on the assumption that the model is powerful enough to ``memorize''\footnote{This may be a poor choice of word as a neural network model always processes the data as a distribution, based on our sufficient statistics point of view.} any number of input data, which is exactly what DynamicEmbedding promises to offer.

Practically, the above argument actually has been confirmed that a reduced number of ``oov'' (out-of-vocab) tokens with a sufficient large dataset can indeed improve the accuracy significantly (\emph{e.g.}, ~\cite{openai_language19}).
In this subsection, we evaluate DynamicEmbedding's ability to process more input/output features, as well as to make more accurate predictions, than its TensorFlow counterparts.

Given that DynamicEmbedding is completely backward compatible with TensorFlow in terms of system performance and model accuracy, it is time to focus on the additional value it offers.
Needless to say, a model with ever-growing capacity can directly help with recall in model prediction: a system that can predict 100M possible outputs is obviously more desirable than one that can only predict 1M.
However, it is also possible that a system with larger capacity also makes less accurate predictions.
In this subsection, we show that larger capacity often guarantees more accurate results.

Theoretically speaking, using a single ``oov'', ``missing'' or ``unk'' token to take care of those out-of-vocab features would unavoidably lead to degrade in model accuracy.
The absence of words in a sentence may dramatically change its meaning, and it is completely possible that different sentences are treated as equal in a language related model (\emph{e.g.}, the Seq2Seq model).
Also a dictionary is often chosen based on frequency rather than importance, and it is known that high frequency words are often less informative, which is why the term frequency-inverse document frequency (TF-IDF) was invented in information retrieval~\cite{info_retrieval_manning}\footnote{This is consistent with the idea proposed by context-aware machine learning~\cite{caml}, in which ``oov'' is redefined as the context-free part of a word/sentence.}.
Besides, our system also provides long-short-term memory components to handle noisy or uninformative inputs.
Therefore, it is safe to abandon the concept of dictionary all together when using DynamicEmbedding.

\begin{figure}[t]
\centering
\begin{tabular}{cc}
    \mbox{\epsfig{figure= 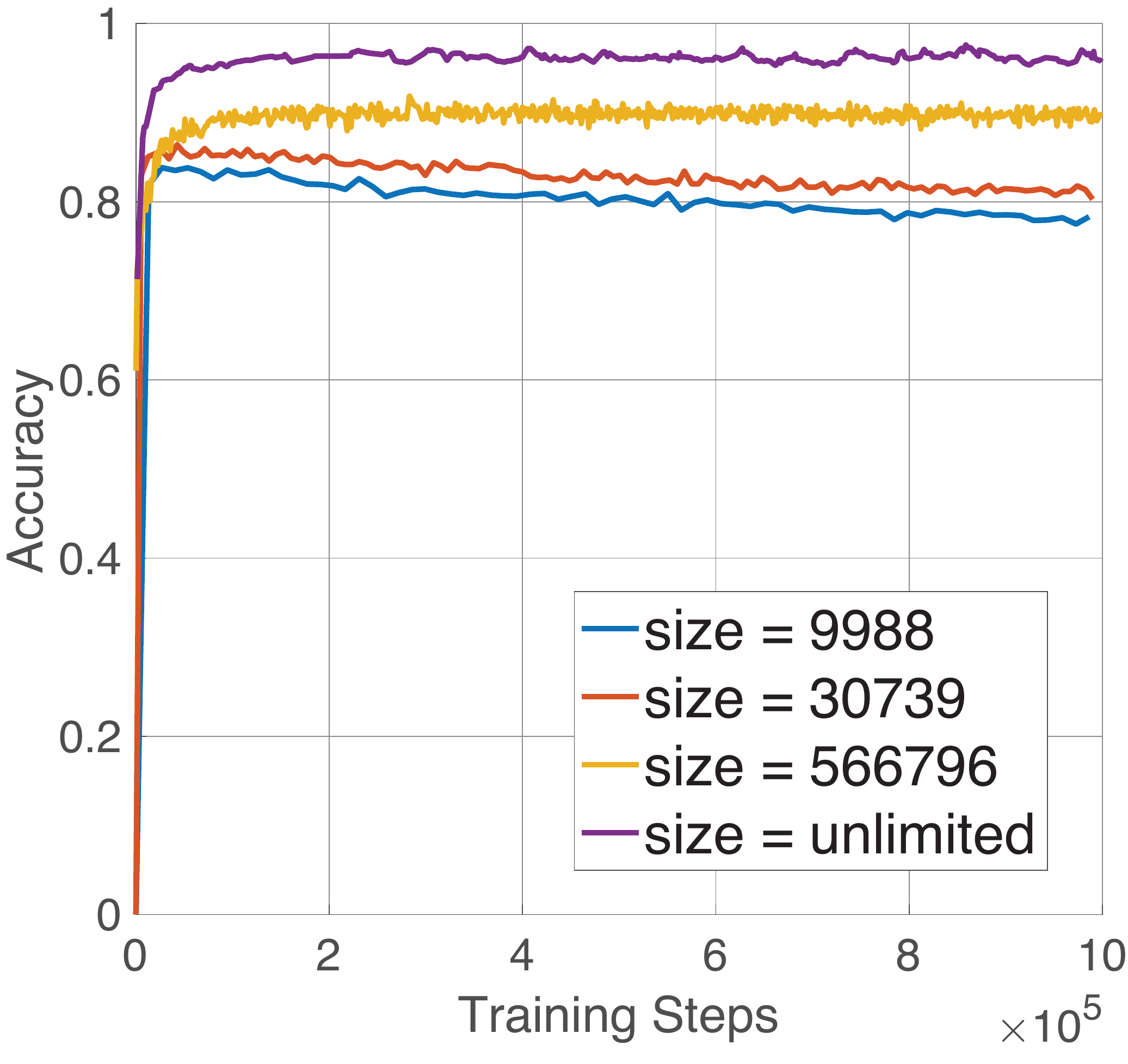, height = 4cm}} &
    \mbox{\epsfig{figure= 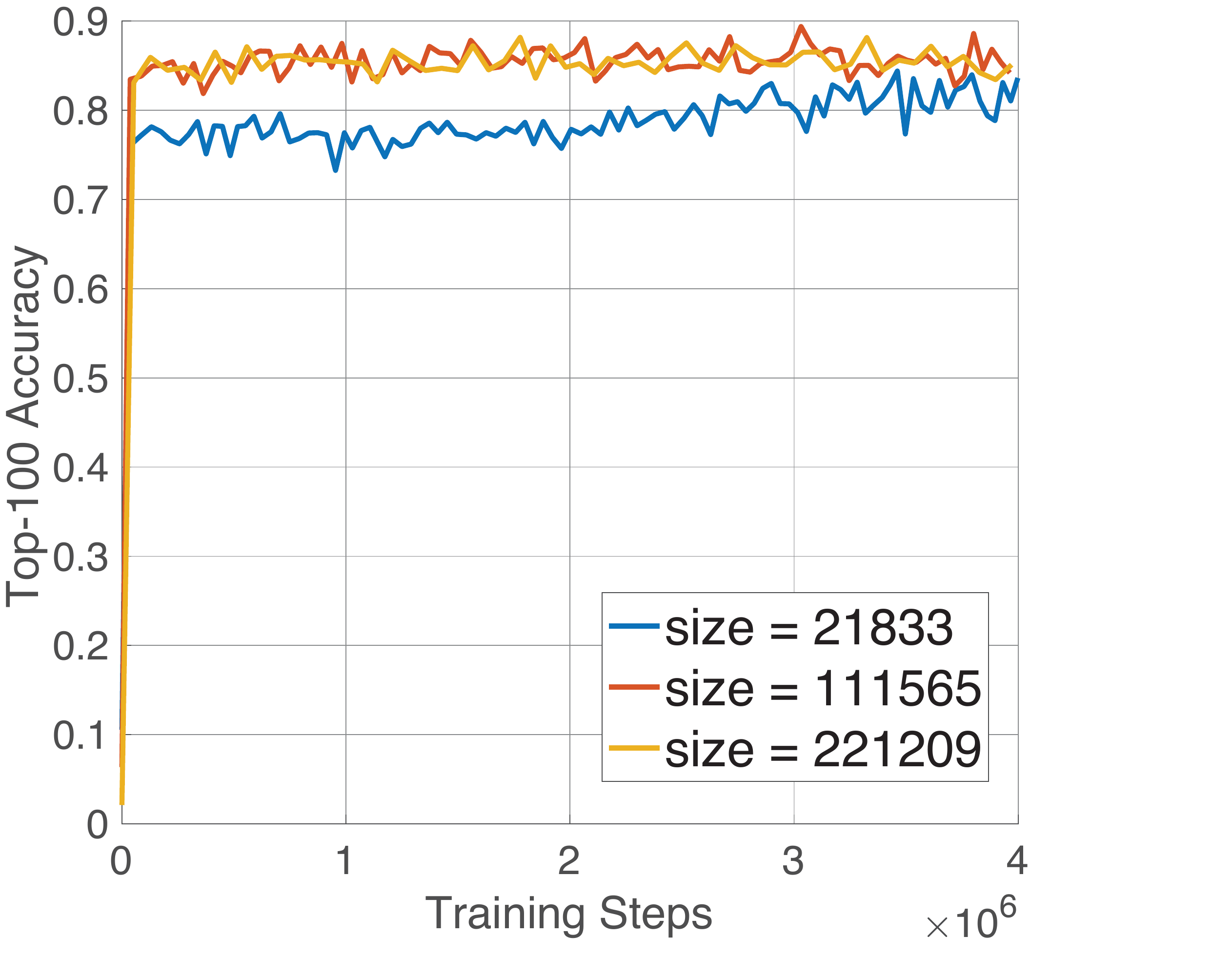, height = 4cm}} \\
    {\scriptsize (a) Word2Vec} & {\scriptsize (c) Sparse2Seq} 
\end{tabular}
\caption{\small (Best viewed in color) The effect of dictionary size on prediction accuracy for two different models.} \label{fig:dict_size}
\end{figure}

Fig.~\ref{fig:dict_size} (a) illustrates the comparison in prediction accuracy between Word2Vec models with and without restrictions in dictionary.
Here we use the same training and test datasets for both models (total file size $\sim$300M), and only change the input dictionary size.
As expected, larger dictionary size consistently lead to better accuracy.

We also tested the impact of dictionary size on the WMT 2016 dataset (total file size $\sim$1.6G) using our Sparse2Seq model as show in Fig.~\ref{fig:dict_size} (b).
The top-100 accuracy for the testing data during training is used for the comparison.
Again, dictionary size is proven to be an important factor for achieving better accuracy, although the improvement seems to be marginal after the size is above a certain threshold.

\subsection{System benchmarks}\label{subsec:system_performance}
Two metrics that are considered important for a learning system are \emph{overall training speed}, measured by global steps per second (GSS) and \emph{resource usage}, measured
by memory usage and CPU rates, etc. 
A large GSS is critical for training on big dataset and fast iteration for model development; a small usage of memory/cpu can guarantee training stability (\emph{e.g.}, less machine failures due to exceeded resources).

\begin{figure}[t]
\centering
\begin{tabular}{ccc}
    \mbox{\epsfig{figure= 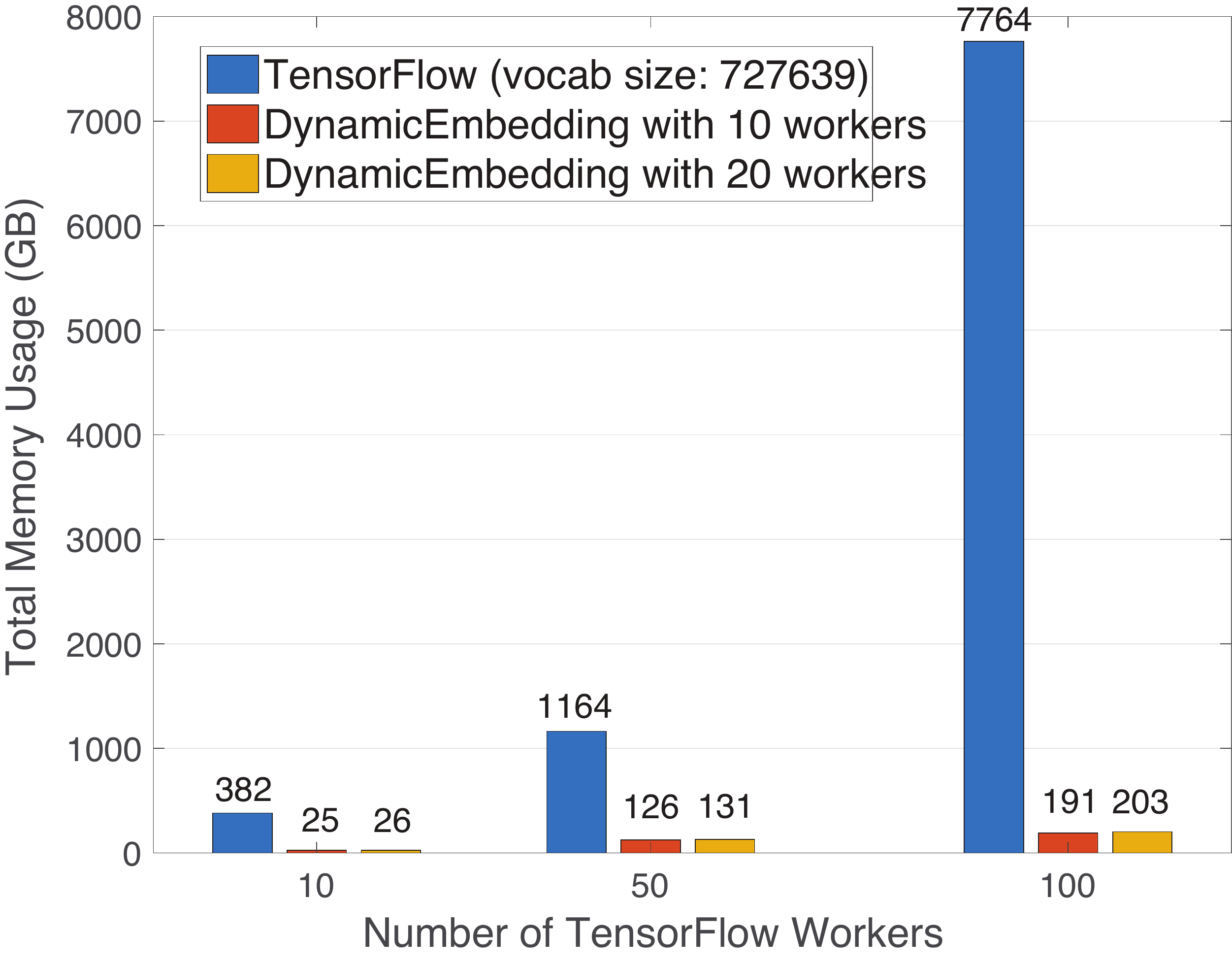, height = 3.4cm}} &
    \mbox{\epsfig{figure= 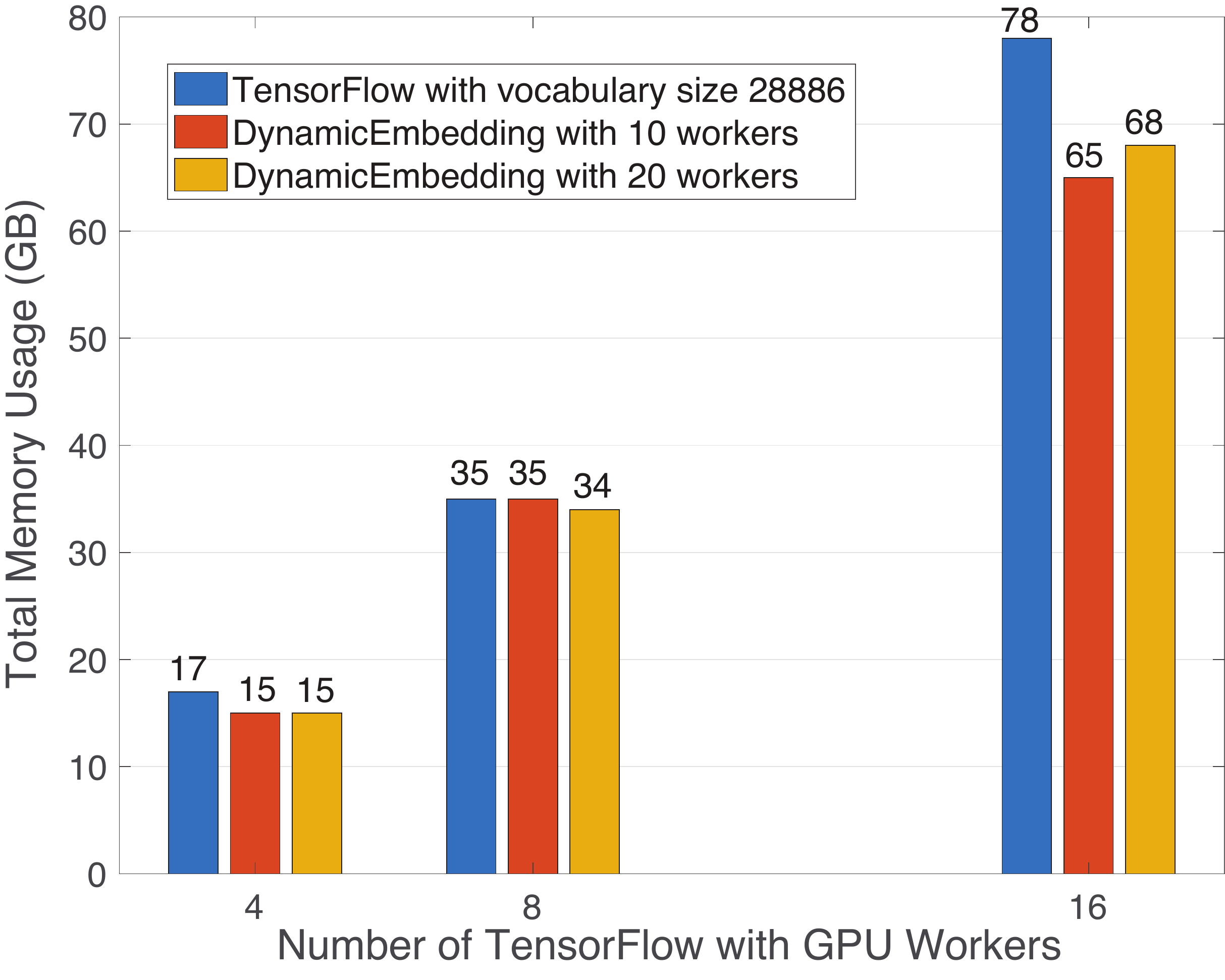, height = 3.4cm}} &
     \mbox{\epsfig{figure= 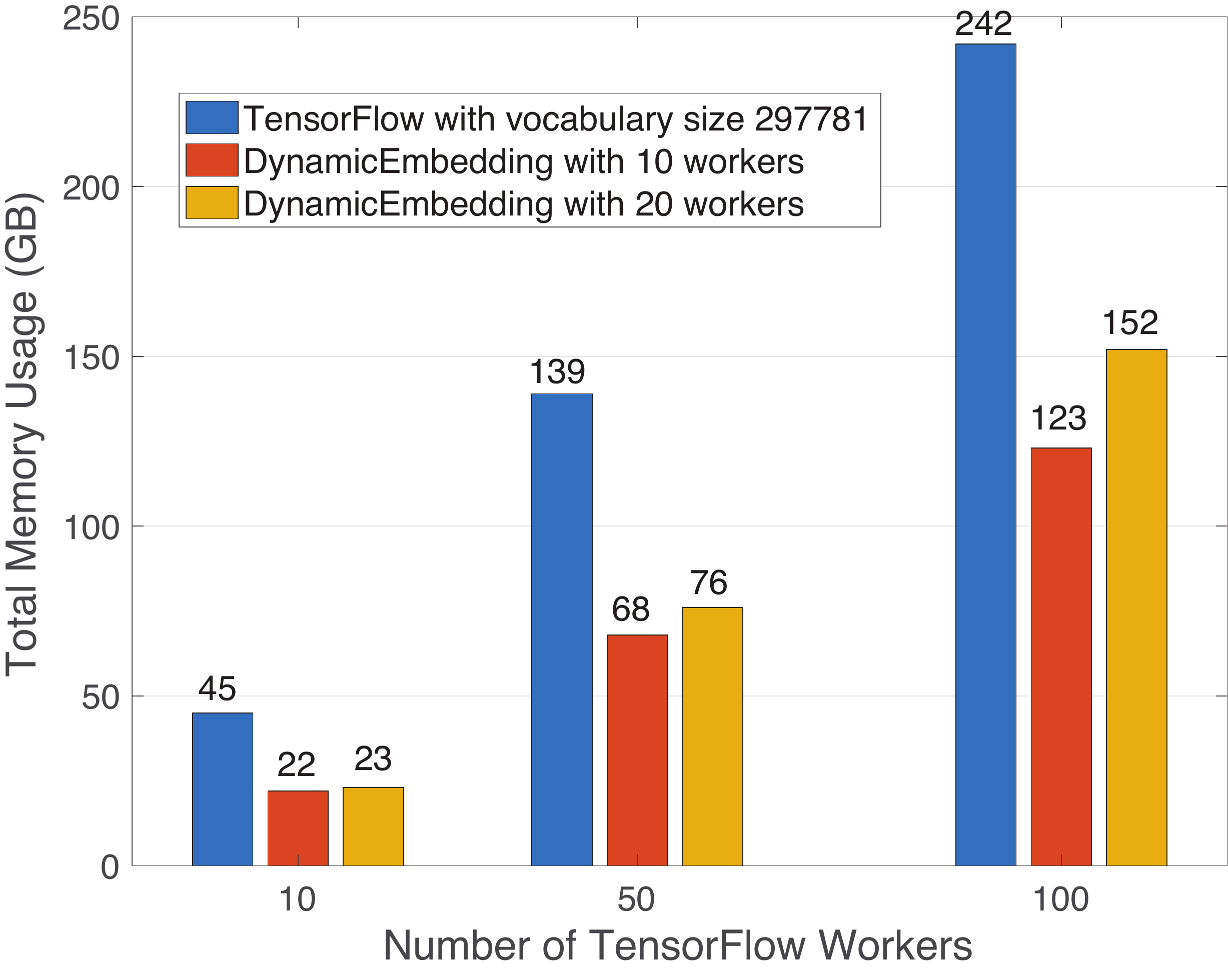, height = 3.4cm}} \\
    {\scriptsize (a)  Word2Vec (DE workers mem.: 30G)} & {\scriptsize (b) Image2Label  (DE workers mem.: 5G)} & {\scriptsize (c) Seq2Seq  (DE workers mem.: 14G)}
\end{tabular}
\caption{\small (Best viewed in color) Comparisons in memory usage between TensorFlow and DynamicEmbedding for three different models. It can be seen that the larger the distinct feature size (or vocabulary size), the more total memory (TensorFlow workers + DynamicEmbedding workers) that can be saved by DynamicEmbedding. Also the total memory consumed by DynamicEmbedding is independent of its number of workers.} \label{fig:memory_perf}
\end{figure}

\begin{figure}[t]
\centering
\begin{tabular}{ccc}
    \mbox{\epsfig{figure= 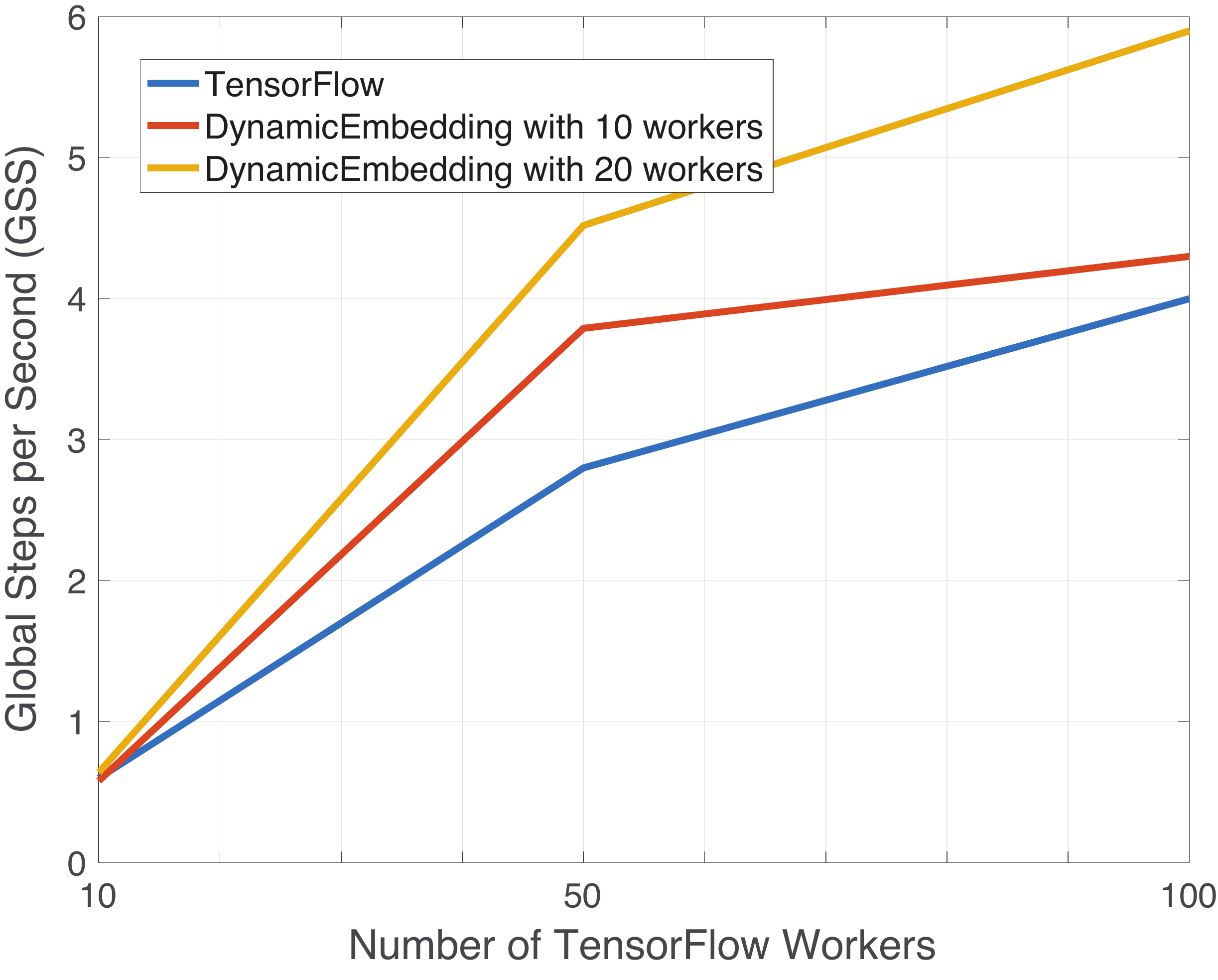, height = 3.4cm}} &
    \mbox{\epsfig{figure= 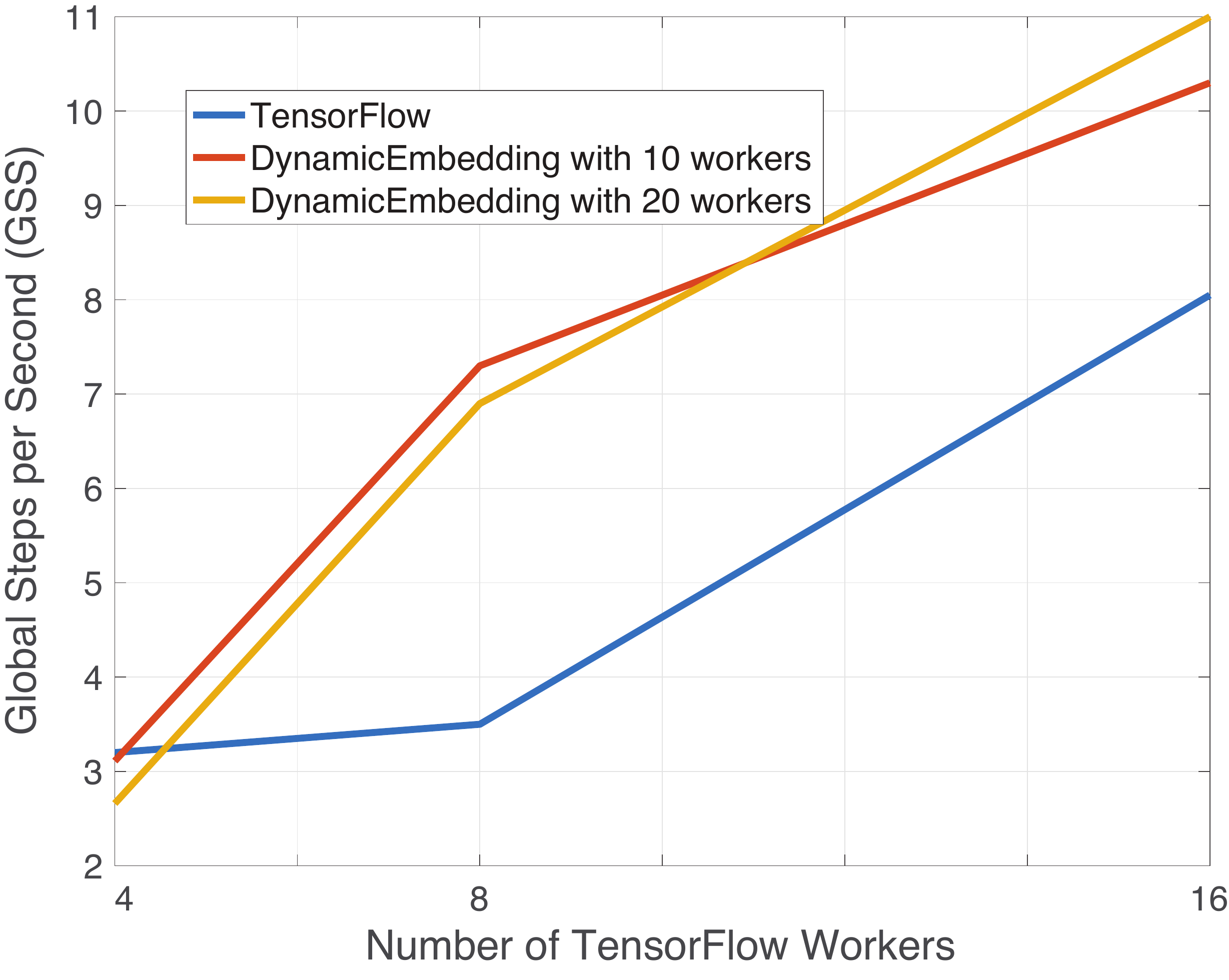, height = 3.4cm}} &
     \mbox{\epsfig{figure= 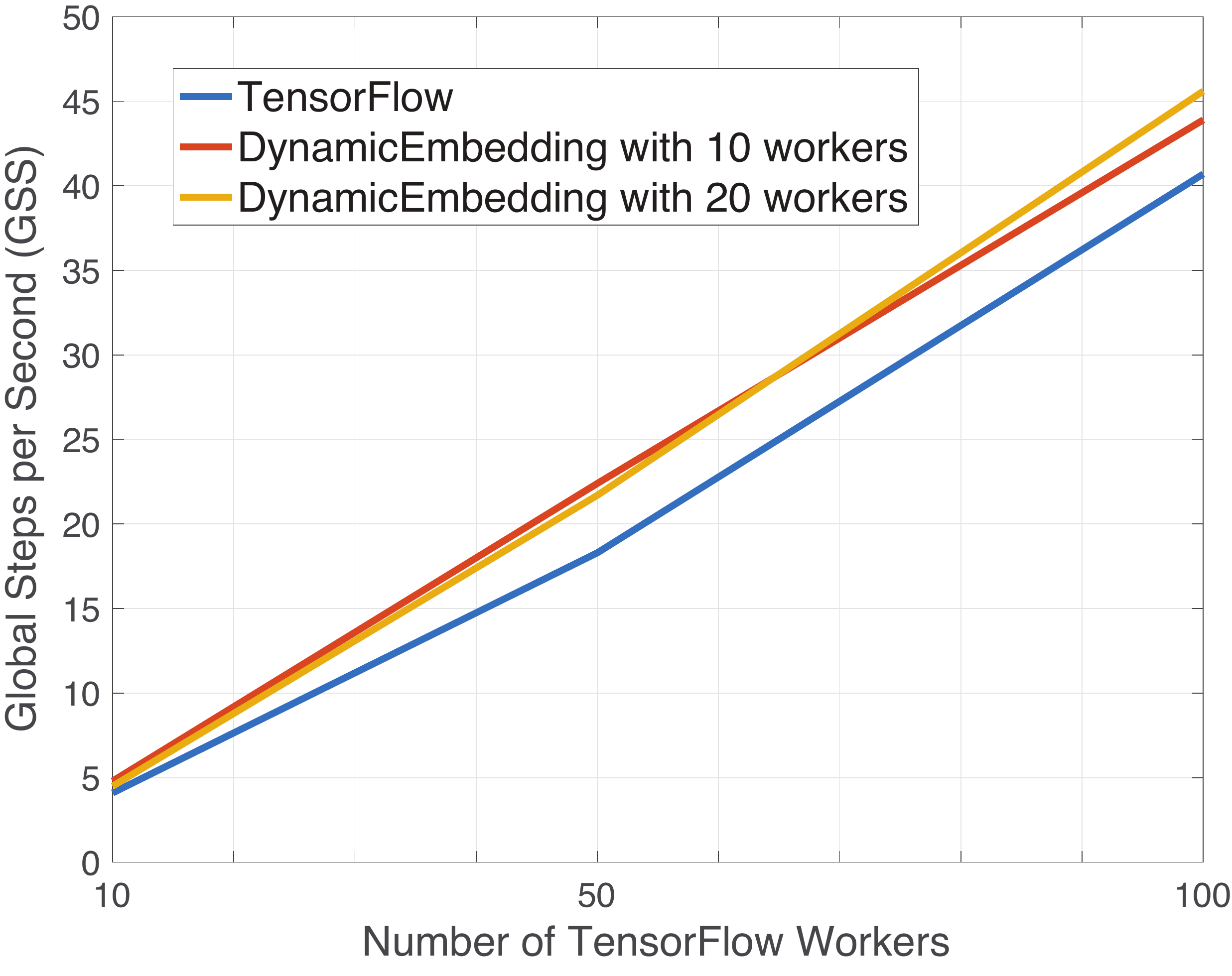, height = 3.4cm}} \\
    {\scriptsize (a)  Word2Vec} & {\scriptsize (b) Image2Label} & {\scriptsize (c) Seq2Seq}
\end{tabular}
\caption{\small (Best viewed in color) Comparisons in training speed between TensorFlow and DynamicEmbedding for three different models. From our experience, both DynamicEmbedding and TensorFlow achieve comparable GSS. DynamicEmbedding is often more stable due to less model loading time and less often worker reschedule since the model sizes for TensorFlow have become much smaller.} \label{fig:gss_perf}
\end{figure}

Our first benchmark dataset consists of cooccurring search query data for the Word2Vec model~\cite{emb_mikolov2013word2vec}.
To allow TensorFlow and DynamicEmbedding be compared side by side, we created a dictionary of size $727$K based on queries' frequency for a conventional Word2Vec model.
For our new DynamicEmbedding based model, we simply feed the data without any dictionary.
Fig.~\ref{fig:memory_perf}(a) and Fig.~\ref{fig:gss_perf}(a) illustrate comparisons on memory usage and GSS between the two models, respectively, which share exactly the same hyperparameters (\emph{e.g.}, batch size is $64$, learning rate is $0.01$ and embedding dimension is $100$) except that DE-Word2Vec does not have a dictionary.
It can been seen that the memory requirement for TensorFlow is marvelously reduced to $1\% - 10\%$ and the memory for DynamicEmbedding workers remains stable regardless of the worker number (close to the total embedding data size, \emph{i.e.}, $\sim 20G$).
Also GSS only depends on the number of TensorFlow workers.
Note that the training dataset is quite challenging for memory as each batch may contain 30M to 50M embedding keys due to large cooccurrence size (many queries may leads to the same website).

Fig.~\ref{fig:memory_perf}(b) and Fig.~\ref{fig:gss_perf}(b) show another benchmark on the Google Inception model~\cite{inception} trained with GPU acceleration.
Our model is trained with up to $16$ machines equipped with NVIDIA Tesla P100 GPU.
In this case, only the output softmax layer involves potentially unlimited number of labels.
Fig.~\ref{fig:memory_perf}(b) and Fig.~\ref{fig:gss_perf}(b) are the performance comparisons between TensorFlow and DynamicEmbedding with different worker sizes. 
We see the same trend as above that GSS only depends on the number of TensorFlow workers, and DynamicEmbedding is running a bit faster than its TensorFlow counterpart, as its candidate sampling is distributed into different DynamicEmbedding workers.

As a more complex example, we re-implemented the full GNMT model~\cite{gnmt} using DynamicEmbedding's APIs.
GNMT is a recurrent neural network model that involves sparse features (words) on both input and output, and the number of features can range from one to hundreds.
Fig.~\ref{fig:memory_perf}(c) and Fig.~\ref{fig:gss_perf}(c) illustrate the performances evaluation, which shows DynamicEmbedding's ability to reduce memory usage while maintaining the same level of GSS.

In sum, DynamicEmbedding is able to dramatically reduce the memory usage of TensorFlow workers when the dictionary size goes up, as it has moved most of the content part of a model into DES.
The total memory usage for the DES only depends on the total size of the embedding, or it can be even lower if only part of the embeddings are loaded into memory (Fig.~\ref{fig:memory_perf}).
Increasing the number of DynamicEmbedding workers only helps parallelism, whereas increasing TensorFlow workers would lead to increase of total memory, as multiple copies of the embedding data are distributed into different workers. 
In terms of training speed measured by GSS, DynamicEmbedding and TensorFlow are quite comparable for gradient descent based optimization (Fig.~\ref{fig:gss_perf}).

\subsection{DE-Sparse2Label model for Google Smart Campaigns}\label{subsec:megabrain}
As DynamicEmbedding is designed to solve real-world problems with extremely large scale, we demonstrate one of our efforts in keyword suggestion for Google Smart Campaign advertisers~\cite{awx}.
Google Smart Campaign is devoted to automating the process of online advertising by smart algorithms powered by AI technologies.
One of the major steps in campaign optimization is to recommend relevant keywords based on the content of advertisers' websites provided during signup\footnote{\url{https://www.google.com/adwords/express/}}.
Although there are billions of $\langle$webpage, keyword$\rangle$ examples from Google Search everyday, not every advertiser's website has been visited by sufficient number of users.
Therefore, our goal is to learn from existing webpage-to-keyword matching examples and generalize it to new websites from our advertisers where allowed.

\begin{figure}[t]
\centering
\begin{tabular}{c}
    \mbox{\epsfig{figure= 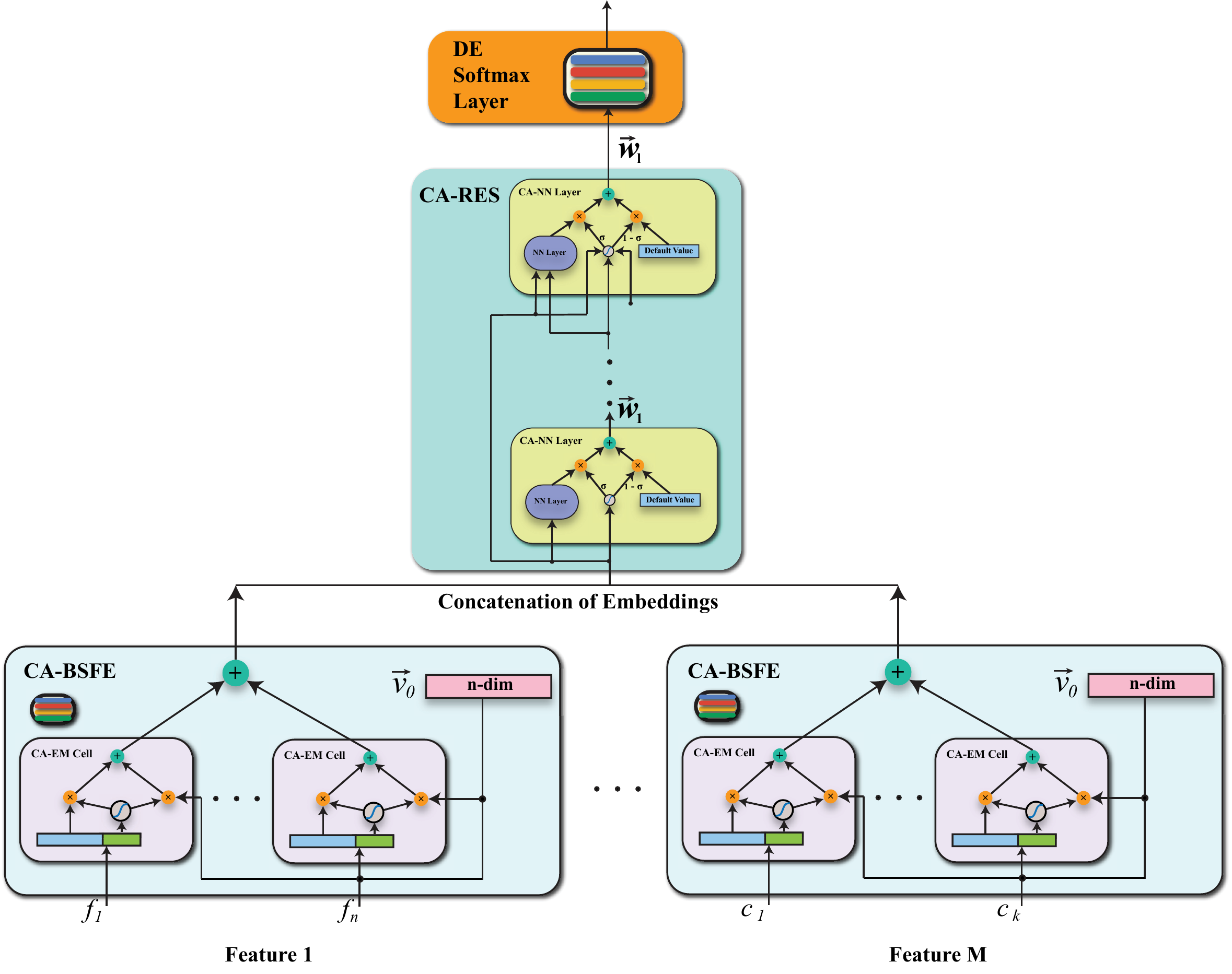, height = 8.2cm}}
\end{tabular}
\caption{\small Architecture of our DE-Sparse2Label model for keyword suggestion. The input to the model are various sparse features from a website. The target for the model is simply a keyword. Both the input and target contain potentially unlimited number of feature values.} \label{fig:megabrain_model}
\end{figure}

Fig.~\ref{fig:megabrain_model} shows the architecture of our keyword suggestion model.
The input to our model is a list of sparse features of a webpage from different annotators to identify its various topics (\emph{e.g.}, freebase entities~\cite{kg_paper}) and representative keywords.
Because each feature can have multiple values, we employ the context-aware CA-BSFE model introduced in~\cite{caml} that composites the embeddings of different sparse feature values based on weights that are related to their context-freeness.
After the embeddings of each feature is computed, they are concatenated and fed into the higher hidden layer, which we employ the ResNet-like CA-RES model~\cite{caml} to further filter out irrelevant information.
At the very top, we use the sampled softmax layer implemented by DynamicEmbedding to handle arbitrary number of output keywords as training target.

One important feature of our model is that it is capable of suggesting keywords for websites in more than $20$ languages that are supported by Google Smart Campaigns.
Previously without DynamicEmbedding, we were only able to launch one model per language with a very limited keyword dictionary size (<1M).
DE-Sparse2Label is the first \emph{all-language} model we have ever trained and it turns out to work remarkably well: ad groups\footnote{\url{https://support.google.com/google-ads/answer/6298?hl=en}} with DynamicEmbedding suggested keywords are able to outperform those non-DynamicEmbedding models in key metrics such as click through rate (CTR), etc  (DynamicEmbedding wins 49 out of a total of 72 revaluation metrics we used for dozens of different countries that we evaluated).
Hence, we have proven that it is no longer necessary to partition the data based on languages, which is a common practice among large-scale applications.

\begin{figure}[t]
\centering
\begin{tabular}{c}
    \mbox{\epsfig{figure= 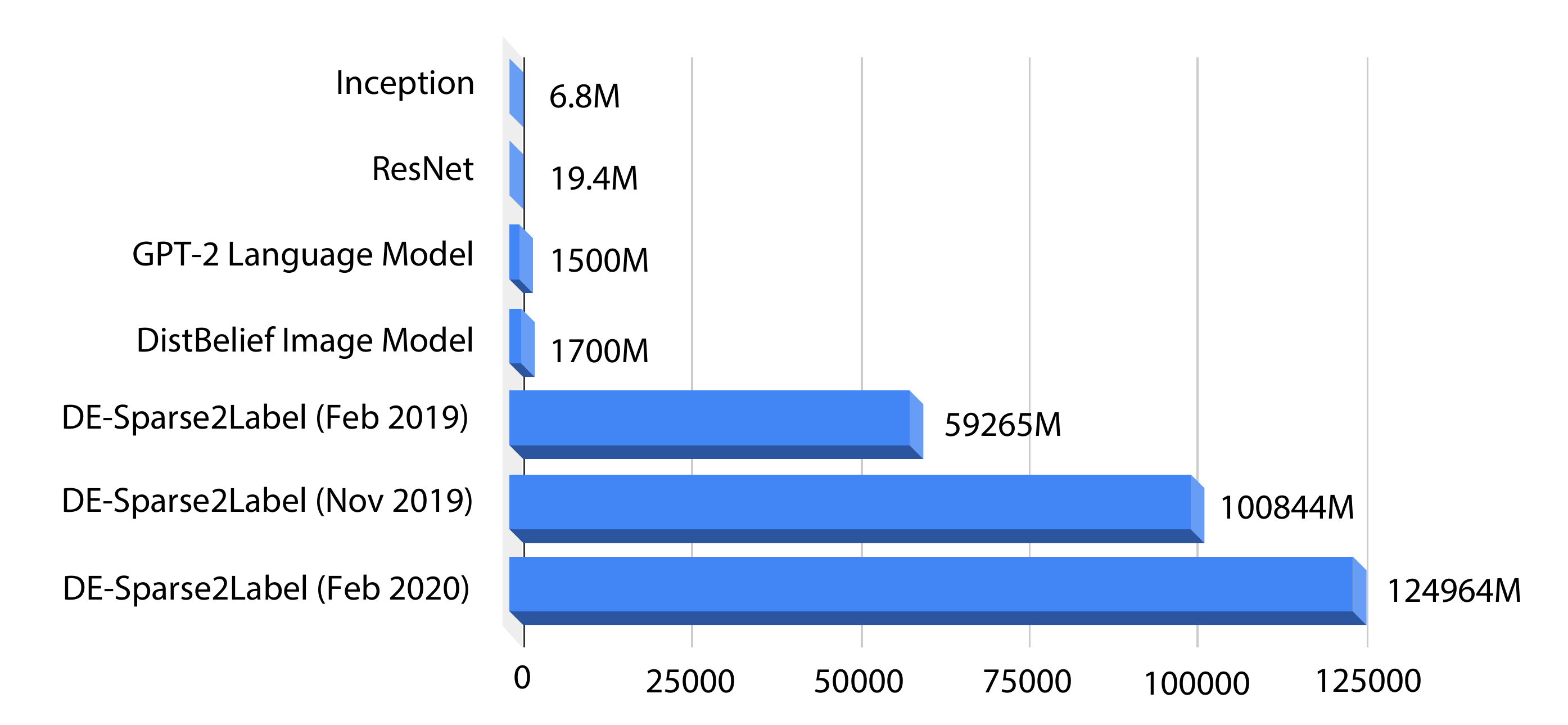, height = 6.2cm}}
\end{tabular}
\caption{\small Comparison in model size among different models, namely Inception~\cite{inception}, ResNet~\cite{resnet_2015}, GPT-2 language model~\cite{openai_language19}, DistBelief image model~\cite{distbelief} and our DE-Sparse2Label model (Fig.~\ref{fig:megabrain_model}).} \label{fig:megabrain_model_size}
\end{figure}

\paragraph{Data and model sizes} Our training data are simply $\langle$website, keywords$\rangle$ pairs from Google Search.
We only select those popular and commercially related ones for training. Our model has been fed with new training data every month and its size (TensorFlow variables + DynamicEmbedding embeddings) has been automatically growing from a few gigabytes to hundreds of gigabytes in less than six months.
Fig.~\ref{fig:megabrain_model_size} compares existing ``large'' models'  parameter sizes with our model at different checkpoints (an $n$-dimensional embedding is considered to have $n$ parameters in the traditional view), and our model has already far exceeded any existing known models in size. 

\subsubsection{Quality evaluation}
One challenge of evaluating the quality of our model suggested keywords is the discrepancy between \emph{keyword} and \emph{query}, as the suggested keywords may not be exactly mapped to Google search queries.
The matching between keywords to queries is controlled by match type~\footnote{\url{https://support.google.com/google- ads/answer/7478529?hl=en}} and by using the default broad match option, our selection of keywords needs to be very specific (\emph{e.g.,} ``wedding'' is considered a bad keyword for a website that does wedding photography).

\begin{figure}[t]
\centering
\begin{tabular}{c}
    \mbox{\epsfig{figure= 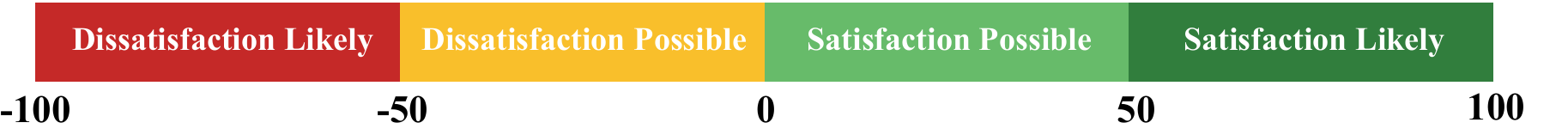, height = 1.0cm}}
\end{tabular}
\caption{\small Keyword quality evaluation design. We ask $5$ human raters to evaluate the extent of search query to website match by giving a score between $[-100, 100]$ and computes their average as the final rating. The final GB ratio of an overall evaluation is computed as the number of examples with score $\geq 50$ (good match) over the number of examples with score $\leq 0$ (bad match). See appendix how each of the four categories are evaluated.} \label{fig:gb_rating}
\end{figure}

To accurately measure the real impact of our suggested keywords to Smart Campaign Advertisers, we employed a very strict metric as follows: after a keyword is served to Google search query, we directly ask 5 different trained human evaluators to give a score between $[-100, 100]$ to rate the match between the search query and advertiser's website (Fig.~\ref{fig:gb_rating}).
The average of the 5 scores is used as the final score for each example.
To evaluate the overall quality of $n$ such query-to-webpage examples (\emph{a.k.a.}, impressions), denoted as $I$, we use the following GB ratio\footnote{This metric was originally proposed by Google Dynamic Search Ads team: \url{https://support.google.com/google-ads/answer/2471185?hl=en}}:
\begin{align}
GB(I) = \frac{|\{i| score(i) \geq 50, i \in I\}|}{|\{i| score(i) \leq 0, i \in I\}|},
\end{align}
where $score(i) \in [-100, 100]$ denotes the average score from 5 different raters for impression $i$.
\emph{Note that our model is completely unaware of such good/bad labeling}.

We run the above mentioned evaluation experiment on 387,151 impressions (\emph{i.e.}, website $\rightarrow$ keyword $\rightarrow$ search query tuples).
Besides the keywords suggested from its softmax output layer, our model also yields a score for each $\langle$webpage, keyword$\rangle$ pair as a confidence of the relatedness of the given keyword to the webpage.
Here the score is computed as the cosine distance between webpage and keyword embeddings computed through our model.
Therefore by using different threshold values (\emph{i.e.}, only serve keywords with keyword-to-webpage scores greater than the given threshold), we can expect to control the quality of the suggested keywords.
Note that our model is able to score any $\langle$webpage, keyword$\rangle$ pairs. 

\begin{figure}[t]
\centering
\begin{tabular}{cc}
    \mbox{\epsfig{figure= 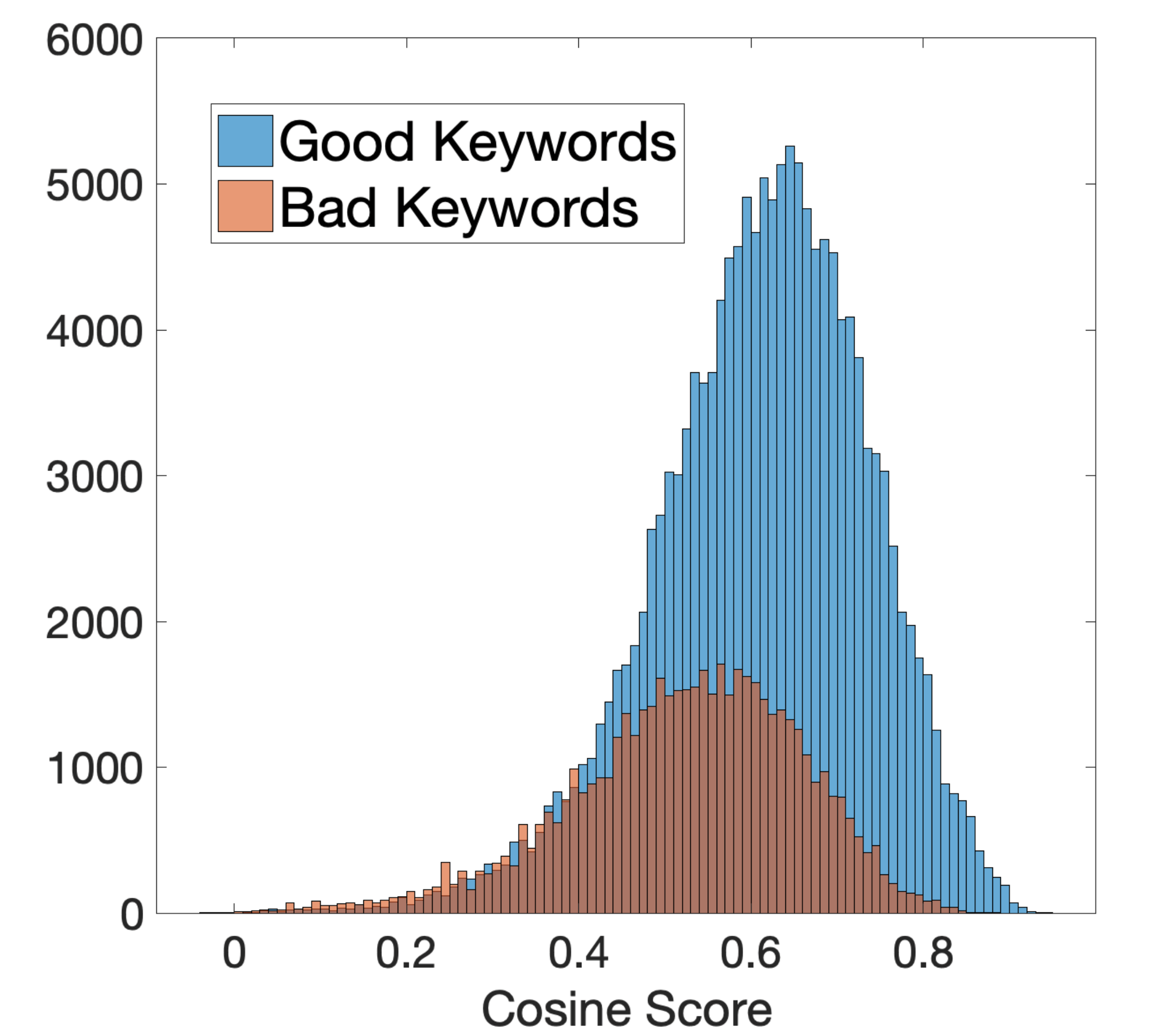, height = 4cm}} &
    \mbox{\epsfig{figure= 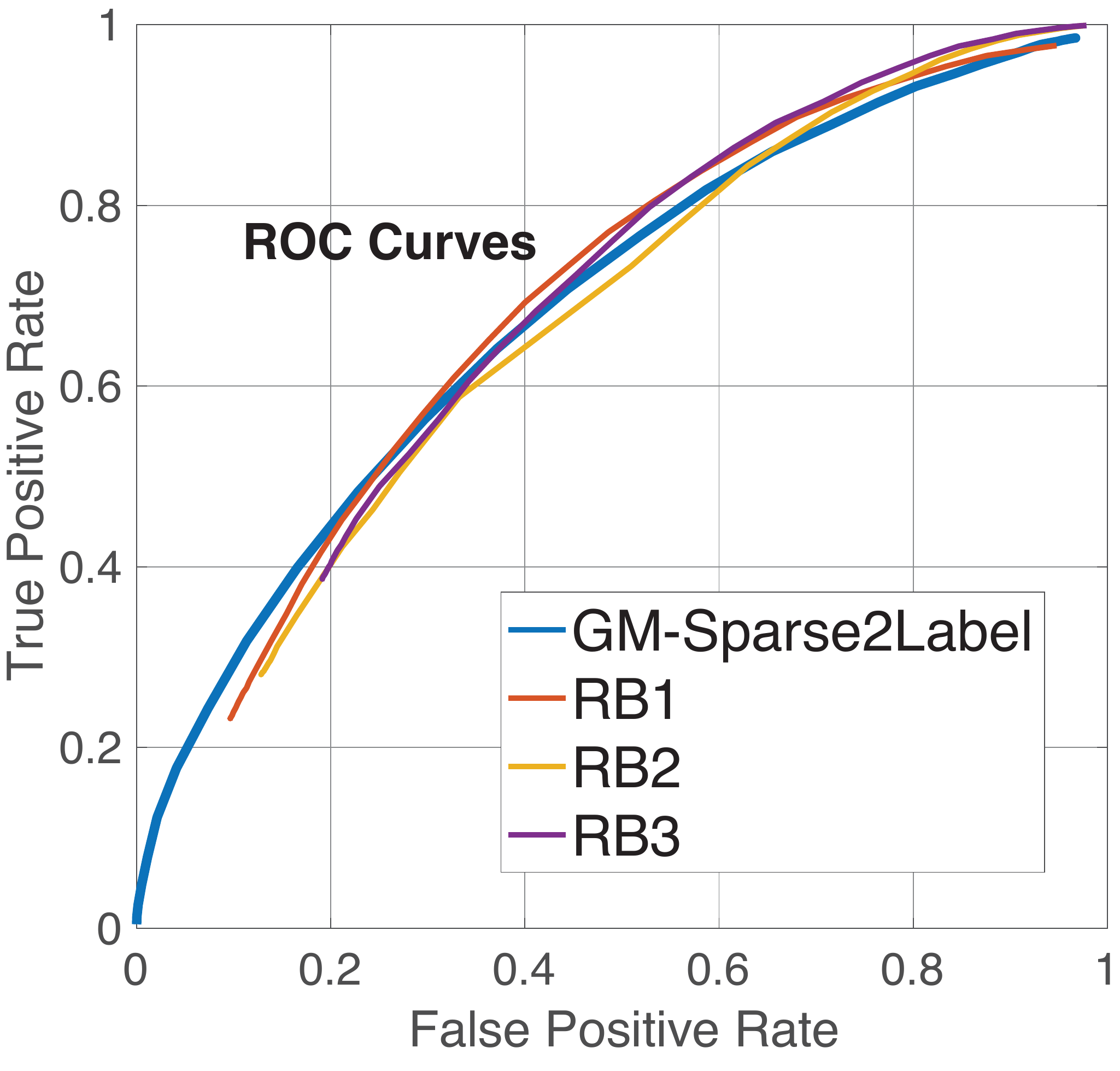, height = 4cm}} \\
    {\scriptsize (a)} & {\scriptsize (c)} 
\end{tabular}
\caption{\small (Best viewed in color) The candidate website to keyword matches contain more good than bad ones and the histograms of the cosine scores from our model indicates that it can identify those good ones quite reliably when the scores are higher than certain threshold (\emph{e.g.}, $0.7$) as shown in (a). In contrast, we compared the GB ratio versus threshold between our model and other rule-based scoring systems available inside Google on the same dataset and (b) shows the ROC curve based on good/bad keywords kept from different thresholds.} \label{fig:megabrain_gb_ratio}
\end{figure}

Fig.~\ref{fig:megabrain_gb_ratio} (a) shows the distribution of the scores for \emph{good} (score $\geq 50$) and \emph{bad} (score $\leq 0$) matches respectively.
Although the two distributions overlap significantly when scores are low (\emph{e.g.}, below 0.6), our model becomes extremely accurate at telling good keywords apart from bad ones when the scores are higher (\emph{e.g.}, above 0.7).
To see this, we compared the GB ratio versus different threshold values between our model and other rule-based scoring algorithms in Fig.~\ref{fig:megabrain_gb_ratio} (b), where we denote these algorithms as RB1, RB2 and RB3, respectively.
We measured the precision/recall tradeoff using the ROC curve based on the total good and bad examples for different threshold values (Fig.~\ref{fig:megabrain_gb_ratio} (b)).
Here the true positive rate for each threshold is computed as the number of kept good keywords over all possible good keywords.
Likewise, the false positive rate is computed as the number of kept bad keywords, above a given threshold, over all possible bad keywords.
It can be seen that our model underperforms traditional rule-based systems slightly under lower thresholds but significantly outperforms others at higher thresholds (when true positive rate becomes smaller).
In practice, this implies that as most of our advertisers need no more than a few thousands keywords, we may be able to guarantee our suggested keywords have extremely high accuracy if our model is used to score all possible keywords. 

Lastly, after our system is fully deployed into production for nearly a year, our post-launch human evaluation on real traffic confirms that the keywords suggested by our model achieves the \emph{highest} GB ratio, as well as the \emph{lowest} bad ratio (percentage of bad keywords over all served keywords), among various keyword sources used in Smart Campaigns.

\section{Conclusion}
\label{section:conclusion}
We demonstrated a preliminary implementation of our new cell model, DynamicCell, has already extended the capacity of TensorFlow significantly through delegation to external storage systems, resulting in a system that clearly outperforms its rule-based counterparts.
We hope that these solutions can be used in a wide variety of ML applications which face challenges around ever-growing scale in data inputs. There may also be future improvements as we look at what we can learn across machine learning, neuroscience and system design etc. Hopefully, by going across these disciplines, we will make faster progresses in AI together.


\subsubsection*{Acknowledgments}
Our work received constructive reviews and active helps from multiple teams inside Google.
First of all, we are deeply indebted to Chao Cai, our team leader, for not only coordinating the communications among all the authors, internal reviewers and approvers, but also proposing numerous edits to improve this paper's readability.
We also thank everyone from Google Adwords Express (a.k.a. Google Smart Campaigns) team, in particular our team founder Xuefu Wang, for building a vibrant working environment for innovation.
Our 2017 summer intern Liming Huang built the first prototype of our DE-Seq2Seq model.
Our 2018 summer interns, Brian Xu and Yu Wang, contributed tremendously in the developments of DE-Image2Label and DE-Sparse2Label models.
Our 2019 summer intern Ruolin Jia applied our DE-Sparse2Seq model to an automated product.
We also thank Google Brain team's Alexandre Passos for providing valuable comments on the comparisons between TensorFlow's Eager mode and our work, Rohan Anil for suggesting us to implement bloom filter for embedding update, and Kai Chen for working relentlessly with us on developing the first version of Sparse2Label model using TensorFlow.
Multiple teams from Google Research provided helpful suggestions at different stages of its development:
Li Zhang, Nicolas Mayoraz and Steffen Rendle provided positive reviews on DynamicEmbedding's initial design, encouraging us to further pursue its implementation;
Dave Dopson, Ruiqi Guo and David Simcha were very responsive in helping us resolve technical issues during the development of TopK component;
Da-Cheng Juan, Chun-Ta Lu and Allan Heydon are actively collaborating with us in expanding the functionalities of DynamicEmbedding for a new semi-supervised learning framework, and are working with Jan Dlabal, Futang Peng and Zhe Li  to enable DynamicEmbedding to work with TPU in their image-related projects.
We would like to give our special thanks to Moshe Lichman and Dan Hurt for their very insightful and timely reviews on the theory and system parts of the paper, respectively, which led to important final touches.
Last but not least, our leadership team Chao Cai, Sunita Verma, Sugato Basu, Sunil Kosalge, Adam Juda, Shivakumar Venkataraman, Jerry Dischler, Samy Bengio and Jeffrey Dean provided valuable comments during our internal review process.

\subsection*{Appendix: Instructions for human evaluation of website to search query match quality}
The instructions below elaborate on how the four categories in Fig.~\ref{fig:gb_rating} are rated.
\begin{itemize}
\item \textbf{Satisfaction Likely}:  To receive this rating, a landing page must offer just what the user looked for. If the user wants car reviews, it should offer car reviews. If the user wants car reviews about a specific model, it should offer car reviews about exactly that model. If the user wants a category of product, the landing page should be devoted to or include that exact category of product. For a Satisfaction Likely rating, what the user is looking for should be apparent with no additional action needed by the user. It is permissible, however, to click on a link to get detailed information.
\item \textbf{Satisfaction Possible}: Use this rating if the page is satisfactory but does not immediately present exactly what the user seeks. If the product or service is for sale on the site, but a search or straightforward navigation is required to find the item, select a rating of Satisfaction Possible rather than Satisfaction Likely. If the site offers a very plausible substitute for a particular product specified in the query, it may receive a rating of Satisfaction Possible or lower. If the query is a search for information, and this information can be found without too much trouble on the advertiser site but is not on the landing page, use Satisfaction Possible.  The one exception here being if the user could have found that same information on the search results page before clicking on the ad. If that is the case, the landing page does not deserve a positive rating. 
\item \textbf{Dissatisfaction Possible}: (i) If the page is marginally related to the query and you think that there's a small chance the user would be interested. (ii) If the page can eventually lead to what the user wants, but only through many clicks or through clicks that lead to an entirely different website. (iii) If the page offers something that you think the user might be interested in, but not what the user was looking for and not especially close to it. For example, if the user is looking for baseball gloves, and the landing page offers athletic socks, there's probably some chance that the user might be interested. However, it's not what the user was looking for, and not all that close to it, so it deserves Dissatisfaction Possible. (iv) If the page can eventually give the user what he or she is looking for, but the process is protracted and difficult.
\item \textbf{Dissatisfaction Likely}: (i) If the page has nothing to do with the query. (ii) If the query is for a product or service, and neither the product/service nor anything close to it can be purchased from the page. (iii) If the query or a word in the query has two meanings, it is clear which meaning is intended by the user, and the advertiser responds to the wrong meaning. For example, [cars 2] refers to a movie. A page for a car dealership is clearly a bad landing page for this query, even if it might be a good result for [car sales]. (iv) If the page looks like a scam, you think users could be harmed by it, or it either attempts to trick the user into downloading something by labeling a download button in a confusing way or tries to download a file without action by the user. (v) If the page loads but is completely unusable (for example, because some content does not load, or page doesn't display properly).  If enough of the page does not load at all (for example, you encounter a 404 error), use the Error Did Not Load flag instead of a rating. (vi) If the page is very bad for any other reason.
\end{itemize}

\vskip 0.2in
\bibliography{de_paper}
\bibliographystyle{plain}

\end{document}